\documentclass[twocolumn]{article}

\usepackage{arxiv}

\usepackage[utf8]{inputenc} % allow utf-8 input
\usepackage[T1]{fontenc}    % use 8-bit T1 fonts
\usepackage[colorlinks,
            linkcolor=red,
            anchorcolor=blue,
            citecolor=blue,
            ]{hyperref}       % hyperlinks
\usepackage{url}            % simple URL typesetting
\usepackage{booktabs}       % professional-quality tables
\usepackage{amsfonts}       % blackboard math symbols
\usepackage{nicefrac}       % compact symbols for 1/2, etc.
\usepackage{microtype}      % microtypography

\usepackage[pdftex]{graphicx} \pdfcompresslevel=9
\graphicspath{ {./figs/} }

\usepackage{lipsum}
\usepackage{amsmath}
\usepackage{amssymb}
\usepackage{pifont}
\usepackage{booktabs}
\usepackage{threeparttable}
\usepackage{multicol}
\usepackage{adjustbox,multirow,array}

\let \bs=\mathbf

\def \path {\mathit{path}}

\let \bs = \boldsymbol

\newcommand{\cmark}{\ding{51}}%
\usepackage{cite}  % comment out for biblatex with backend=biber

\title{A Survey on Computational Solutions for Reconstructing Complete Objects by Reassembling Their Fractured Parts}

\author{
 Jiaxin Lu$^\dag$ \\
  Computer Science Department\\
  The University of Texas at Austin\\
  2317 Speedway, Austin, Texas \\
  \texttt{lujiaxin@utexas.edu} \\
   \And
 Yongqing Liang$^\dag$ \\
  Department of Computer Science and Engineering\\
  Texas A\&M University\\
  435 Nagle St, College Station, Texas \\
  \texttt{lyq@tamu.edu} \\
  \And
 Huijun Han$^\dag$ \\
  Department of Computer Science and Engineering\\
  Texas A\&M University \\
  435 Nagle St, College Station, Texas \\
  \texttt{hazelhan@tamu.edu} \\
\And
 Jiacheng Hua$^\dag$ \\
  Department of Computer Science and Technology\\
  Tsinghua University\\
  30 Shuangqing Road, Beijing \\
  \texttt{hjc21@mails.tsinghua.edu.cn} \\
  \AND
  Junfeng Jiang$^*$ \\
  College of Artificial Intelligence and Automation \\
  Hohai University\\
  1915 Hohai Avenue, Changzhou, Jiangsu \\
  \texttt{jiangjf@hhu.edu.cn} \\
  \And
  Xin Li$^*$ \\
  Section of Visual Computing \& Computational Media, and\\
  Department of Computer Science and Engineering \\
  Texas A\&M University \\
  789 Ross Street, College Station, Texas\\
  \texttt{xinli@tamu.edu} \\
  \And
  Qixing Huang$^*$ \\
  Computer Science Department \\
  The University of Texas at Austin \\
  2317 Speedway, Austin, Texas \\
  \texttt{huangqx@cs.utexas.edu} \\
}

\begin{document}
\maketitle
\newcommand\blfootnote[1]{%
\begingroup
\renewcommand\thefootnote{}\footnote{#1}%
\addtocounter{footnote}{-1}%
\endgroup
}
\blfootnote{
$^\dag$ Authors contributed equally.\ $^*$ Corresponding authors.}
\footrule

\onecolumn
\begin{abstract}
   Reconstructing a complete object from its parts is a fundamental problem in many scientific domains. The purpose of this article is to provide a systematic survey on this topic. This reassembly problem requires understanding the attributes of individual pieces and establishing matches between different pieces. Many approaches also model priors of the underlying complete object. Existing approaches are tightly connected problems of shape segmentation, shape matching, and learning shape priors. We provide existing algorithms in this context and emphasize their similarities and differences to general-purpose approaches. We also survey the trends from early procedural approaches to more recent deep learning approaches. In addition to algorithms, this survey will also describe existing datasets, open-source software packages, and applications. To the best of our knowledge, this is the first comprehensive survey on this topic in computer graphics.
\end{abstract}
% keywords can be removed
\keywords{Computing methodologies\and Computer graphics\and Shape modeling\and Shape analysis}

\twocolumn
\section{Introduction}

Reconstructing a complete object from its parts is a fundamental problem in many scientific domains. In archaeology, we collect fragments of one or multiple objects at an archaeological site, and the goal is to restore the original objects. In medical applications, we need to study how to assemble bone fragments to plan bone reduction surgery. This problem is also crucial in paleontology for reconstructing fossils and in forensics for evidence reconstruction. The challenge extends to 3D object/scene modeling, where we want to construct a complete object/scene from parts/objects. Other relevant problems are protein docking and human-object interactions, in which the difference is that the transformation of each part becomes non-rigid. The purpose of this article is to provide a systematic survey on this topic, including algorithms, datasets, open source packages, and applications. To the best of our knowledge, this is the first systematic survey on this topic. 

The reassembly problem is related to many shape analysis tasks. An important task is to segment and classify the parts' surface into fractured and intact regions, which is a special shape segmentation problem. Another important task is to find correspondences between fractured regions of different pieces to find relative poses between them. This is related to scan matching, with the difference that the correspondences are between complementary points. In addition, many approaches utilize priors of the underlying complete object, including symmetries and shape priors. This calls for approaches to analyze and encode such priors and use them to solve the reassembly problem.

Similar to other problems in computer graphics, we have also witnessed the transition from early procedural approaches to recent deep learning approaches. Early approaches have explored all aspects of the reassembly problem, including the segmentation of fractured regions and original regions, matching of fractured regions, and encoding of priors (symmetries and shape spaces) of the underlying complete shape. However, they are restricted to hand-crafted features, so entire systems do not generalize robustly. In addition, it is difficult to encode shape priors among objects with large geometrical and topological variations. These issues can be addressed using deep learning approaches. In addition to the fact that the shape segmentation and shape matching problems become more robust, we also have approaches that perform shape segmentation and shape matching together, which benefit from end-to-end training in deep learning. In addition, with advances in learning 3D shape generative models, it becomes possible to apply learned diverse shapes prior to reassembly.

\begin{figure*}
% \vspace{3in}
\includegraphics[width=\linewidth]{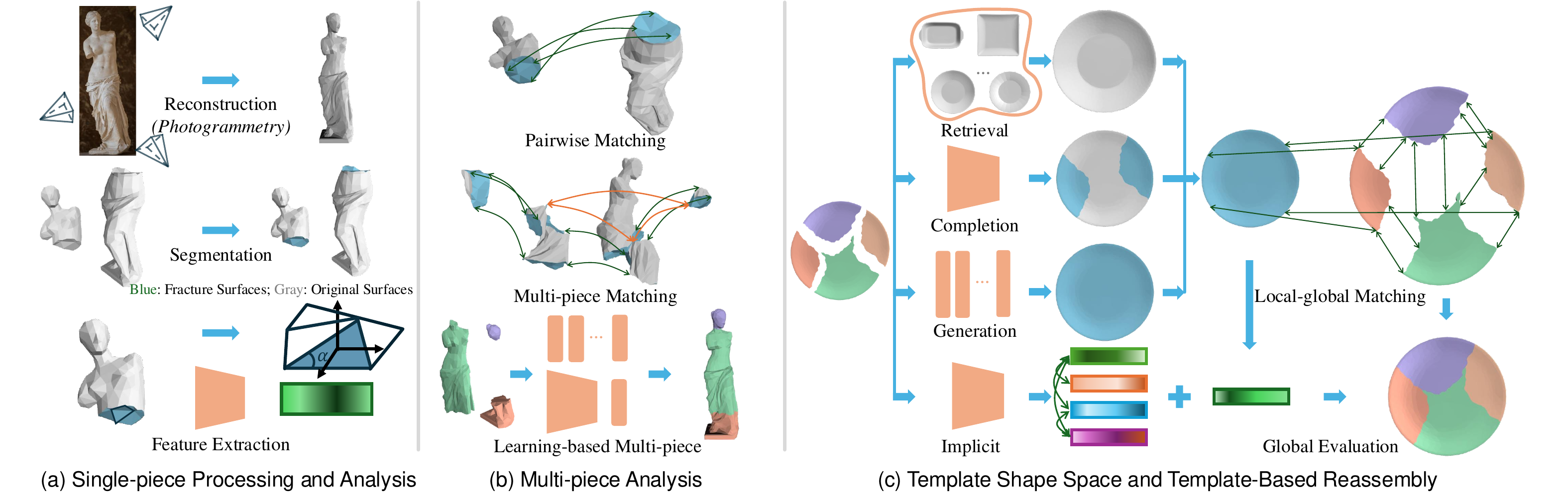}
\caption{Overview of the methods presented in this survey. (a) Single-piece processing and analysis. (b) Multi-piece analysis. (c) Template shape space and template-based reassembly.}
\label{Figure:Overview}
\vspace{-0.1in}
\end{figure*}

Despite tight connections to relevant problems in shape analysis, applying existing approaches to the 3D reassembly problem is not straightforward. This is the case for both procedural and deep learning based approaches. While reassembly shares many technical components with other geometry processing tasks, it has not been formally recognized as a distinct sub-area within geometry processing and shape analysis. Through this survey, we aim to systematically organize state-of-the-art results in this domain and establish a structured foundation for this important research direction.

\subsection{Related Surveys}

The most relevant survey is geometric analysis in cultural heritage~\cite{DBLP:journals/cgf/PintusPYWGR16}, in which reassembling fractured objects is a central topic. The difference between our survey and~\cite{DBLP:journals/cgf/PintusPYWGR16} is two-fold. First, in contrast to their survey which focuses mostly on procedural techniques, our survey covers recent advances using deep learning. Second, in contrast to the specific application of assembling fractured objects, this survey covers many applications, including bone fracture reduction, shape modeling by assembling parts, and applications in biology, geoscience, and paleontology. 

Another relevant survey is~\cite{10.1111:cgf.142660}, which studies how to decompose a shape into rigid parts. Its focus is different in the sense that it focuses on optimizing decompositions with respect to different geometric and physical criteria for manufacturing and fabrication. In contrast, we focus on solving the inverse problem of reconstructing complete objects from parts. \cite{Gigilashvili2023reconarchtextile} discussed techniques for reconstructing fractured archaeological textiles. It focuses more on using textiles as guidance for puzzle solving and archaeological reassembling in both 2D and 3D cases. Our work demonstrates more general cases and applications where textiles might not exist.

Our survey also discusses geometry processing techniques, including 3D reconstruction~\cite{DBLP:journals/cgf/BernardiniR01,DBLP:journals/cgf/BergerTSAGLSS17}, mesh segmentation~\cite{DBLP:journals/cgf/Shamir08,DBLP:journals/cgf/RodriguesMG18}, primitive segmentation~\cite{DBLP:journals/cgf/KaiserZB19}, shape descriptors~\cite{DBLP:journals/cgf/RostamiBRY19} and  correspondences~\cite{DBLP:journals/cgf/KaickZHC11}, symmetry detection~\cite{DBLP:journals/cgf/MitraPWC13}, and data-driven shape analysis and processing~\cite{DBLP:journals/cgf/XuKHK17}. Although the process of reassembling parts uses these techniques, this article focuses on approaches tailored for this particular class of applications, which usually have modified inputs, outputs, and problem setups.

Many reassembly techniques utilize shape priors of the underlying complete priors. This is an area in which we have seen great progress in adopting machine learning models to learn shape priors from data under suitable 3D representations. Recent surveys~\cite{10.1111:cgf.14020,10.1111:cgf.15061,10.1111:cgf.15063} provide excellent coverage of state-of-the-art results. In contrast, this survey focuses on how to use such parametric shape priors to solve the reassembly problem. 

There are also other relevant review articles. \cite{DBLP:journals/jocch/PapaioannouSAMG17} survey related approaches for each step in the pipeline that starts from reassembly to object completion, which also addresses missing regions. \cite{2015:Pottery:Survey} present a survey article on the specific problem of reassembly of pottery fragments. \cite{DBLP:conf/vast/GregorSPSAM14} survey related work on reassembly of defective cultural heritage objects. Our survey presents a comprehensive review of general fractured object reassembly and connections to relevant research results in geometry processing.

%When dealing with fractured parts, data representation often takes the form of meshes or point clouds, essentially graphical structures. Consequently, graph learning emerges as a pivotal tool in this context. For a meticulous exploration of graph learning techniques, refer to \cite{9416834}.

\subsection{Paper Organization}

We organize the remainder of this paper as follows. In Section~\ref{Section:Overview}, we provide a more detailed overview of this paper. In Section~\ref{Section:Single:Piece:Analysis}, we describe existing approaches for analyzing individual pieces. In Section~\ref{Section:Multi:Piece:Analysis}, we discuss existing approaches for analyzing multiple pieces, focusing on problems of matching pairs of pieces and multiple pieces jointly. In Section~\ref{Section:Template:Analysis}, we introduce existing methods for encoding prior knowledge of the underlying complete object. In Section~\ref{Section:Template:Based}, we review approaches that utilize priors about the complete object to solve the reassembly problem. In Section~\ref{Section:Applications}, we discuss various applications of the reassembly problem. Section~\ref{Section:Datasets:Software:Packages} presents available datasets and software packages. In Section~\ref{Section:Future:Directions}, we present current trends and future directions on this topic. Finally, we conclude this survey in Section~\ref{Section:Conclusions}.

% \end{multicols}
% \begin{multicols}{2}
\begin{figure*}[!ht]
    \centering
    \includegraphics[width=\linewidth]{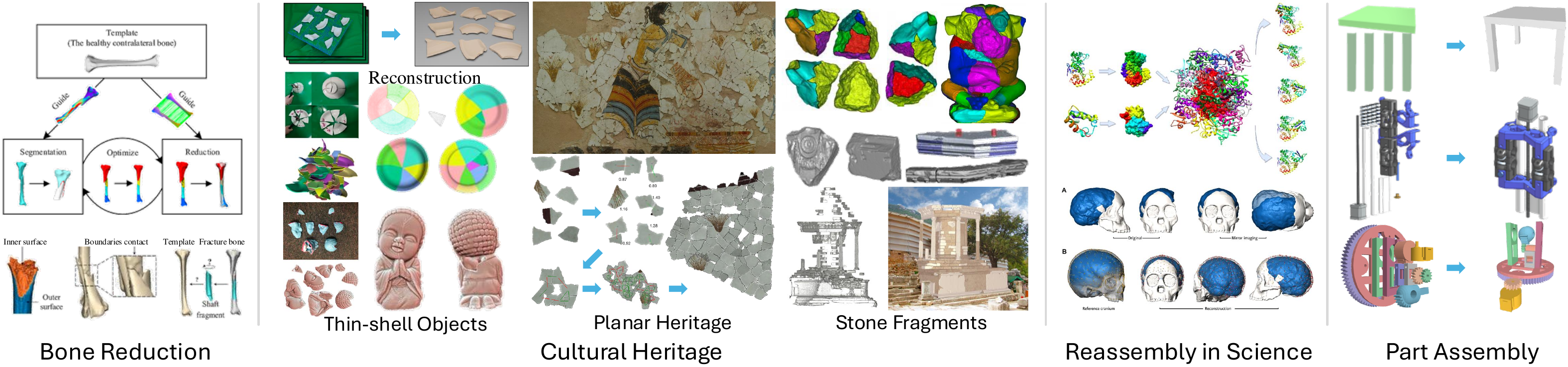}
    \vspace{-0.1in}
    \caption{Overview of applications presented in this survey.}
    \label{Figure:Overview_application}
    \vspace{-0.1in}
\end{figure*}

\section{Overview}
\label{Section:Overview}

This section presents an overview of this survey (See Figure~\ref{Figure:Overview} and Figure~\ref{Figure:Overview_application}). 

\noindent\textbf{Single piece processing and analysis (Sec.~\ref{Section:Single:Piece:Analysis}).} Single piece processing and analysis concern individual pieces of fractured objects. Topics include how to digitize physical fragments into digital 3D models and how to segment each fragment into regions of original surfaces and fractured surfaces. Although these topics are central topics in the field of geometry processing and shape analysis, we focus on techniques that are tailored to the particular characteristics of fragments of physical objects. We will cover both procedural approaches and more recent deep learning approaches.

\noindent\textbf{Multi-piece analysis (Sec.~\ref{Section:Multi:Piece:Analysis}).} Multi-piece analysis studies the critical problem of establishing correspondences between fragments to match them. We begin with pairwise matching techniques using geometric and textural features extracted from the fracture surfaces and original surfaces. We will survey global techniques that do not require prior knowledge of the relative pose between a pair of fragments and local techniques that start from reasonable initial relative poses. We then proceed to discuss methods that involve more than two pieces, in which there is a cycle-consistency constraint among the relative poses along a cycle of pieces. In addition to fragment matching, we will also discuss approaches that solve the correlated problems of fragment segmentation and fragment matching together. Similarly to single-piece analysis, we will cover both procedural approaches and deep learning approaches. 

\noindent\textbf{Template shape space (Sec.~\ref{Section:Template:Analysis}).} Template analysis studies how to encode prior knowledge of the underlying complete object. We focus on techniques that learn the shape spaces of complete objects from the data. Early approaches of this type focus on learning linear shape spaces. Recently, we have seen great progress in adopting generative modeling techniques to learn more complex shape spaces. Challenges on how to apply the learned shape priors for fractured object reassembly are discussed.  

\noindent\textbf{Template-based reassembly (Sec.~\ref{Section:Template:Based}).} We then discuss approaches that used priors of the underlying complete objects to solve the reassembly problem. These include approaches that enforce priors of symmetries and primitive structures and those that optimize a complete object in the learned shape space. In the latter case, a central problem is to establish correspondences between points on the original surfaces of the fragments and points on the template surface.

\noindent\textbf{Applications (Sec.~\ref{Section:Applications}).} After discussing technical approaches to solving the reassembly problem, we will discuss various applications of the reassembly problem. We will discuss five areas of fractured object reassembly application, including bone reduction, restoring thin-shell, planar, and stone fragments, and reassembly problems in science. Finally, we discuss the relevant problem of assembling parts to form complete objects. 

\noindent\textbf{Datasets and open source (Sec.~\ref{Section:Datasets:Software:Packages}).} We discuss publicly available datasets and open source packages. Datasets include early small-scale datasets and recent large-scale real- and synthetic datasets. 

\noindent\textbf{Future directions (Sec.~\ref{Section:Future:Directions}).} The last section of this survey summarizes existing approaches and discusses future directions. We will emphasize how to advance existing reassembly problems by developing better learning approaches, focusing on challenges in representations, data, and utilizing foundation models. Moreover, we will also discuss potential new applications of reassembly in natural and life sciences. 
% \end{multicols}

% \begin{multicols}{2}
\section{Single Piece Processing and Analysis}
\label{Section:Single:Piece:Analysis}

This section presents techniques for single-piece analysis. We begin by examining relevant work on digitizing fragment shapes in Section~\ref{Subsec:Reconstruction}. We then discuss prior work on segmenting a piece into fracture surfaces and original surfaces in Section~\ref{Subsec:Segmentation}. 

%Finally, We discuss approaches that extract features from a given piece in Section~\ref{Subsec:Feature:Extraction}.
\begin{table*}[h!t]
    \caption{Category of scanning technology.}
    \label{tab:scanning}
    \centering
    \begin{tabular}{c|l|l|l|l}
        \toprule
        Categories &  Accuracy  & Efficiency &Cost(\$) & Key features\\ 
        \midrule
        Laser & $0.01\sim3$~mm  & several minites & $500$ - $5$k & Fast capture of large pieces\\
        % \midrule
        Photogarmmetry &    $<=0.15$ mm    & minutes to hours& $200$ - $5$k & Capture of color and texture information \\
        % \midrule
        Structured light &    $0.1$ mm   &  seconds to minutes&$50$ - $5$k & High efficiency and precision\\
        % \midrule
        CT &  $0.001\sim0.5$ mm     &    minutes to hours& $1$k - $10$k & Internal structure detail  capture\\
        \bottomrule
    \end{tabular}
    % \vspace{-0.1in}
\end{table*}

\subsection{Reconstruction}
\label{Subsec:Reconstruction}

Converting physical fragments into digital 3D models involves two key stages: digitization through various scanning technologies and registration to create complete 3D models. Although scanning choices influence data quality, the registration process shares fundamental algorithmic similarities with the assembly task itself.
This section discusses how fragmented pieces can be captured using sensors, each of which requires specific reconstruction techniques. 

\begin{figure}
\centering
\includegraphics[width=\linewidth]{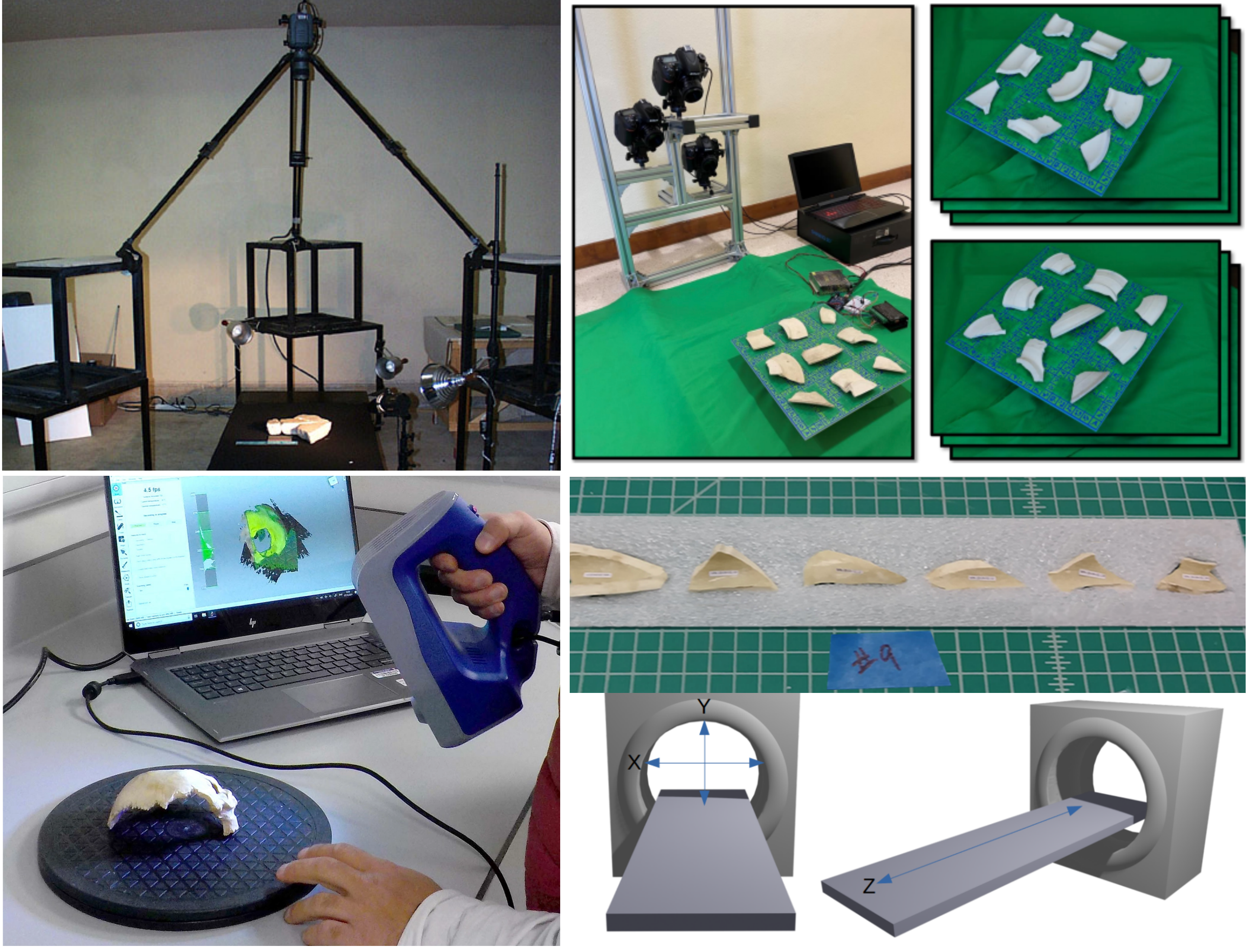}
\caption{Four types of sensors for digitizing fragment shapes. (Top-left) Laser scanning~\cite{10.1145/3596711.3596733}. (Top-right) Photogrammetry capturing~\cite{wang2023batch}. (Bottom-left) Structured light capturing~\cite{GoalTech2024}. (Bottom-right) CT Scanning~\cite{yezziwoodley2022batchartifactscanningprotocol}.}
\vspace{-0.2in}
\label{Figure:Sensors}
\end{figure}

\noindent\textbf{Scanning Technologies.} Several technologies are available for fragment digitization (Figure~\ref{Figure:Sensors}):

\emph{Laser Scanning} captures high-resolution geometry ($\sim0.01$ mm) by measuring reflected laser beams. High-frequency systems like HandySCAN ($1.8M$ points/second) enable rapid capture~\cite{bouzakis20163d,rudari2024accuracy}, while lower-frequency options like NextEngine provide more affordable alternatives. Laser scanning excels in outdoor environments and has enabled major digitization efforts like Stanford's ``Forma Urbis Romae'' dataset~\cite{10.1145/3596711.3596733} and the ``Reconstructing Thera Frescoes'' project~\cite{10.1145/1360612.1360683}. However, the lack of color capture and high equipment costs are notable limitations.

\emph{Photogrammetry} achieves resolution comparable to that of laser scanning ($\sim0.15$ mm)~\cite{wyatt2022after, wang2023batch} using a regular camera. State-of-the-art multi-view stereo algorithms achieve pixel-level accuracy for 90\% of reconstructed points~\cite{5226635}. This approach is particularly suited for field work due to its low cost and minimal equipment requirements. Porter et al.~\cite{porter2014portable} designed a specialized rig for field collection without electricity, requiring approximately 20 minutes and 100 photographs per artifact~\cite{bleed2017photogrammetrical}. Although sensitive to lighting conditions and require controlled illumination~\cite{Magnani-2014}, recent protocols have demonstrated a high throughput of 730 fragments per day~\cite{wang2023batch}. However, photogrammetry struggles with certain materials such as water, glass, and steel~\cite{koutsoudis2015structure}, and requires careful selection of focal length~\cite{BISSONLARRIVEE2022e00224}.

\emph{Structured Light} balances speed and accuracy ($\sim0.1$-$0.2$ mm) by projecting patterns onto objects~\cite{ruiz2021comparative}. It can process 60 pottery fragments per hour~\cite{KARASIK20081148} or 5-6 reflective artifacts per hour~\cite{GROSMAN20083101}. While capturing both geometry and color, it works best indoors and may struggle with shiny or dark surfaces. Notable applications include the work by Fan et al.~\cite{10.1145/2980179.2980225}.

\emph{CT Scanning} is uniquely capable of capturing internal structures~\cite{kelley2018digital,lin2010application} at micron-scale resolution ($1$-$500$ microns)~\cite{abel_digital_2011}. It has been applied successfully to rocks~\cite{RAMANDI2017817}, stones~\cite{goldner2022practical}, fossils~\cite{lautenschlager2016reconstructing}, bones~\cite{lautenschlager2014cranial}, and teeth~\cite{olejniczak2006assessment}. Recent work demonstrates batch processing of 220 artifacts in two hours at 140-micron resolution~\cite{goldner2022practical}. Important datasets include skull scans~\cite{mangrulkar2021automated} and detailed anatomical structures~\cite{brown2014restoration}.

\noindent\textbf{Registration and 3D reconstruction.} 
In order to obtain the complete geometry of an object, it is necessary to scan from multiple angles and then use point cloud registration to align these scans in a common coordinate system.  
When reference markers are possible, such as small disc-shaped stickers \cite{bouzakis20163d} or unevenly spaced grid lines drawn on a turntable \cite{Magnani-2014}, they facilitate point cloud registration by serving as common reference points.  
When using such markers is infeasible, markerless registration such as the Iterative Closest Point (ICP) algorithm and its variants are utilized to align the point clouds~\cite{KARASIK20081148}. 
A good initial alignment is essential for the ICP algorithm, as it directly affects how quickly and accurately the algorithm can find the final alignment~\cite{7368945}.
Therefore, careful manual annotations are essential for optimum outcomes. 
In order to improve registration efficiency, \cite{wang2023batch} devised an automated and robust registration technique that leverages the bilateral boundary ICP, offering reliable initial relative pose estimates to initiate the ICP.
After forming the dense point cloud, a mesh modeling algorithm, such as marching cubes~\cite{lorensen1998marching} or Poisson reconstruction~\cite{kazhdan2006poisson} is used to generate the surface.
Unlike traditional methods that separate scanning and surface reconstruction, \cite{qixing2007bayesian} combines these steps into a single process to create cleaner and more complete 3D models. Using joint registration, the method significantly improves reconstruction accuracy, ensuring better consistency when combining different views. The experimental results indicate that it can perfectly reconstruct the surfaces of archaeological fragments. This approach was used to reconstruct the fragments of the Octagon project~\cite{thuswaldner2009digital}.

Batch processing in scanning improves efficiency and reduces the need for manual intervention. This approach is widely adopted in large-scale applications.
\cite{10.1145/2980179.2980225} developed an automated system that can scan multiple 3D objects by intelligently planning camera views and scanning paths, reducing the need for human input. It enhances scanning efficiency and quality with structured light scanning and calibrated positioning, making it ideal for large-scale applications like archaeological documentation. 
\cite{10.1145/1360612.1360683} introduced a batch processing system that achieves efficient scanning through parallel operation of multiple devices, automated processing, and a fast matching algorithm. A single operator can manage up to four 3D scanners, while another uses a flatbed scanner to capture high-resolution images and normal maps. The system automatically aligns and processes data in the background, minimizing manual intervention. The fast geometry-based matching further optimizes efficiency, allowing the system to process approximately 10 fragments per hour. \cite{wang2023batch} presented another batch system to scan two sides of fragments and developed custom registration and reconstruction techniques to process the paired scans.

\subsection{Segmentation}
\label{Subsec:Segmentation}

Surface segmentation is a common preprocessing step in the analysis of individual fragments for reassembly tasks, aimed at breaking down complex surface geometries into simpler, meaningful regions. Typically, this involves distinguishing between the original and fracture surfaces, facilitating easier analysis and subsequent matching with other fragments to achieve successful reassembly.

\begin{figure}
% \vspace{2in}
\includegraphics[width=\linewidth]{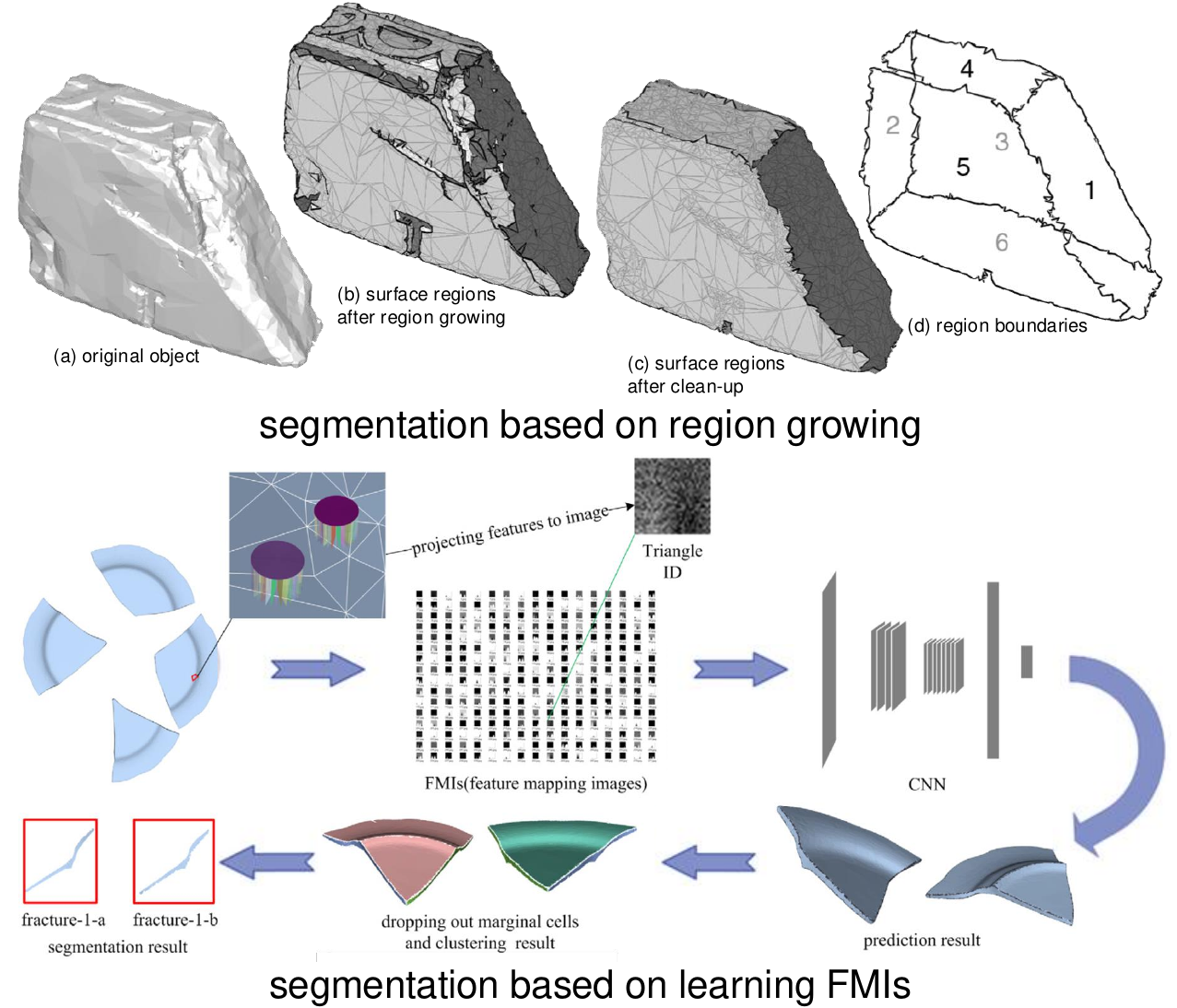}
\caption{(Top) Procedural methods for fracture surface segmentation \cite{905491}. (Bottom) Deep learning methods for fracture surface segmentation \cite{LIU2021102963}.}
\label{Figure:Segmentation:Single:Piece}    
\vspace{-0.2in}
\end{figure}

\noindent\textbf{Geometry-based methods.} Geometry-based surface segmentation methods rely on the intrinsic geometric properties of the fragment's surface, such as curvature, edges, and other spatial features. These methods are among the most established techniques in the field due to their reliance on fundamental geometric principles (see Figure~\ref{Figure:Segmentation:Single:Piece} (Top)). Several geometry-based methods have been developed for surface segmentation. For example, \cite{905491} employs a simple region-growing algorithm based primarily on normal vectors, along with a facet segmentation threshold, to initially segment the original object into distinct surface regions. Following this segmentation, bumpiness is used to detect fractured surfaces. \cite{10.1145/1141911.1141925} first perform multi-scale edge extraction on point-sampled surfaces, which is constrained to return closed cycles to obtain initial face cycles. These face cycles are then partitioned into original and fracture surfaces using an iterative normalized cut method on a weighted graph, constructed considering surface sharpness and roughness. \cite{:10.2312/egp.20141060} use a hierarchical agglomerative clustering algorithm that starts with each element as its own cluster, merging the closest clusters based on a user-defined metric until no further merges are possible. A custom 2-level caching scheme is used to reduce redundant distance calculations and sorting, enhancing efficiency. \cite{10.1016/j.jcde.2014.12.002} first apply the Laplace operator to smooth the surface, and then use a clustering algorithm based on a vertex normal vector to acquire rough surface segmentation, remove noise surfaces, and then refine the segmentation by merging adjacent faces using surface roughness and face normal vectors. In addition to these, various other clustering and graph segmentation algorithms, such as K-means clustering \cite{LuoCoseg2013}, hierarchical clustering \cite{YAN20121072}, mean-shift clustering \cite{1563229}, spectral clustering \cite{6701205} and random walks \cite{10.1145/3414685.3417806}, can also be adapted for fractured surface segmentation. These techniques leverage explicit geometric rules and properties, making them interpretable and efficient with limited data. However, their performance may decrease when handling complex surfaces or noisy input.

\begin{figure*}
% \vspace{3in}
\includegraphics[width=\linewidth]{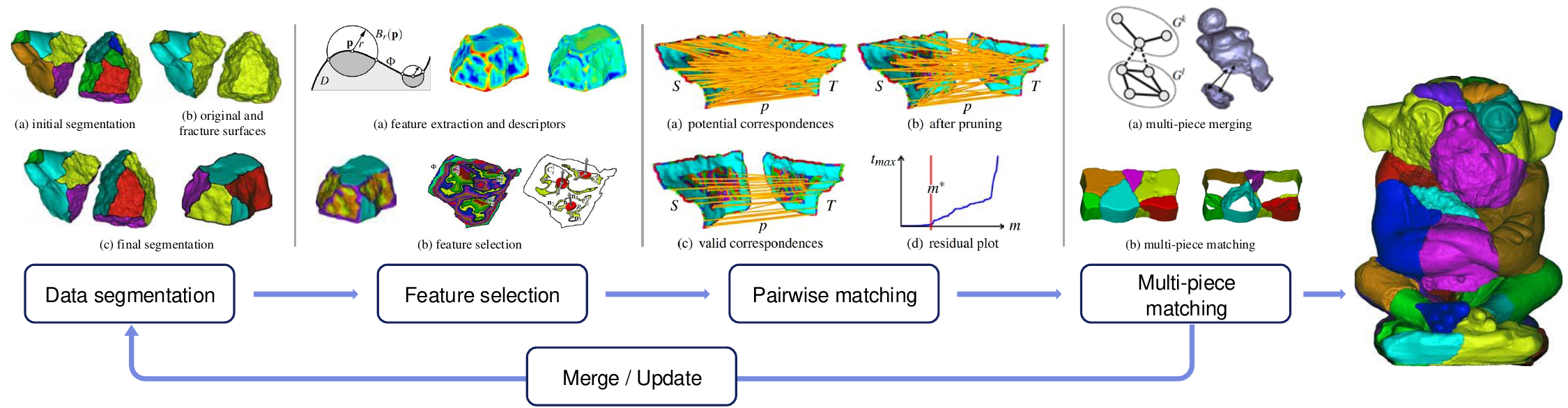}
\caption{A representative multi-piece shape analysis framework presented in~\cite{10.1145/1141911.1141925}. Pair-wise matching use patch features extracted from the fracture surface and segment features extracted from fracture surface boundaries. A voting-based approach is used to find consistent correspondences between geometric features. Matching among pieces employs a greedy approach. }
\label{Figure:Non-Deep Learning:Matching}    
% \vspace{-0.1in}
\end{figure*}

\noindent\textbf{Learning-based methods.} Learning-based surface segmentation techniques have emerged as a promising direction, as they can automatically discover patterns and features from data without requiring hand-crafted geometric rules. These methods offer new possibilities for surface segmentation in fragment reassembly tasks. Although some learning-based fractured object reassembly methods take the entire fragments as input for analysis, bypassing prior surface segmentation, there are still methods that decouple the reassembly pipeline, retaining segmentation to achieve better matching and reduce computational costs. Learning-based segmentation methods are predominantly based on graph learning techniques. The initial segmentation process can be conceptualized as a binary classification task. For example, \cite{LIU2021102963} utilize a convolutional neural network (CNN) to learn the labels of the ridge and non-ridge cells (see Figure~\ref{Figure:Segmentation:Single:Piece}(Bottom)), facilitating the extraction of the fracture margins of the fragments. Subsequently, fractured surfaces can be well segmented by trimming fragment models along these margins. \cite{lu2023jigsaw} employ Multi-Layer Perceptron (MLP) layers combined with a sigmoid activation function to predict confidence in the identification of a fracture point. If we generalize segmenting fractured surfaces to 3D semantic segmentation, we can also leverage advanced general-purpose graph learning algorithms like PointNet \cite{Qi-2017-CVPR}, PointNet++ \cite{NIPS2017-d8bf84be} and DGCNN \cite{dgcnn} to perform segmentation. These powerful frameworks offer robust solutions for intricate 3D data analysis, and they will also be mentioned in the next section. These frameworks provide effective tools for analyzing complex 3D data, as we will discuss further in the next section. While these approaches can potentially handle challenging cases with complex surfaces and noise, they typically require large training datasets and substantial computational resources.

\section{Multi-Piece Analysis}
\label{Section:Multi:Piece:Analysis}

This section reviews the literature on multi-piece analysis, which considers two or more fragments. In Section~\ref{Subsec:Pairwise:Matching}, we review pairwise matching techniques. In Section~\ref{Subsec:Multi:Piece:Matching}, we discuss multi-piece matching techniques, including early optimization-based techniques and recent techniques based on deep pose regression. Finally, we discuss the correlated problem of joint segmentation and matching in Section~\ref{Subsec:Joint:Seg:Matching}.

\subsection{Pairwise Matching}
\label{Subsec:Pairwise:Matching}

\noindent\textbf{Rigid scan matching.} A relevant problem of pairwise piece matching is rigid scan matching, which finds the relative rigid pose between two depth scans. Given two point sets $P$ and $Q$, rigid scan matching aims to find a rotation $R \in SO(3)$ and translation $t \in \mathbb{R}^3$ that minimize:
\begin{equation}
\min_{R,t} \sum_{i=1}^{n} \min_{j} ||Rp_i + t - q_j||^2
\end{equation}
where $p_i \in P$ and $q_j \in Q$ are points from the respective scans. A standard pipeline of rigid scan matching consists of feature extraction, feature matching, and rigid pose fitting. Early approaches use hand-crafted geometric descriptors, e.g., SpinImage~\cite{DBLP:journals/pami/JohnsonH99} and Integral invariant~\cite{DBLP:journals/cagd/PottmannWHY09} features. These features usually only encode local information; therefore, the resulting feature matches have many false positives. The rigid pose fitting procedure addresses this issue by enforcing the rigidity constraint, which finds a subset of feature matches that share a rigid transformation. Popular rigid pose fitting approaches include RANSAC~\cite{DBLP:journals/cgf/SchnabelWK07}, general Hough transform~\cite{DBLP:journals/tog/MitraGP06,DBLP:journals/tog/PaulyMWPG08,DBLP:journals/cgf/MelladoAM14}, and their variants~\cite{DBLP:journals/tog/AigerMC08,DBLP:journals/cgf/BouazizTP13}. Another category of methods formulates rigid scan matching as spectral matching~\cite{DBLP:conf/iccv/LeordeanuH05}, which computes a consistency matrix among candidate feature matches. The elements of this consistency matrix encode whether distances and angles are preserved between the corresponding pairs of correspondences. Rigid pose fitting is formulated as computing a sub-matrix in which all elements are consistent, and is done using spectral techniques. 

More recent approaches develop deep learning techniques for rigid scan matching. These approaches employ deep neural networks to extract features and develop differentiable layers to fit rigid poses. Examples include differentiable RANSAC and attention-based rigid regression. Representative methods in this category include DCP~\cite{DBLP:conf/iccv/LuWZFYS19}, PNetLK~\cite{DBLP:conf/cvpr/AokiGSL19},
FCGF~\cite{DBLP:conf/iccv/ChoyPK19},
Predator~\cite{DBLP:conf/cvpr/HuangGUWS21}, RoReg~\cite{DBLP:journals/pami/WangLHWCDGWY23}, and GeoTF~\cite{DBLP:journals/pami/QinYWGPIHX23}. These approaches demonstrate promising results on benchmark datasets with similar training and testing conditions. Developing approaches that generalize well to diverse real-world scenarios remains an important research direction.

\noindent\textbf{Procedural fragment matching.} Similar approaches have been developed for rigid fragment matching, which also combines feature extraction, feature matching, and rigid pose fitting. However, rigid fragment matching and rigid scan matching have several notable differences. These differences come from several aspects, including where to extract features and what constraints are induced by feature matching. Another important difference is the robustness of the extracted features, the feature descriptors, and the matching procedure, as the surfaces of the fragments may be severely damaged or there are tiny fragments between two fragments, causing geometric differences. Figure~\ref{Figure:Non-Deep Learning:Matching} shows a representative framework for fragment matching.

\begin{figure}[t]
% \vspace{2in}
\includegraphics[width=\linewidth]{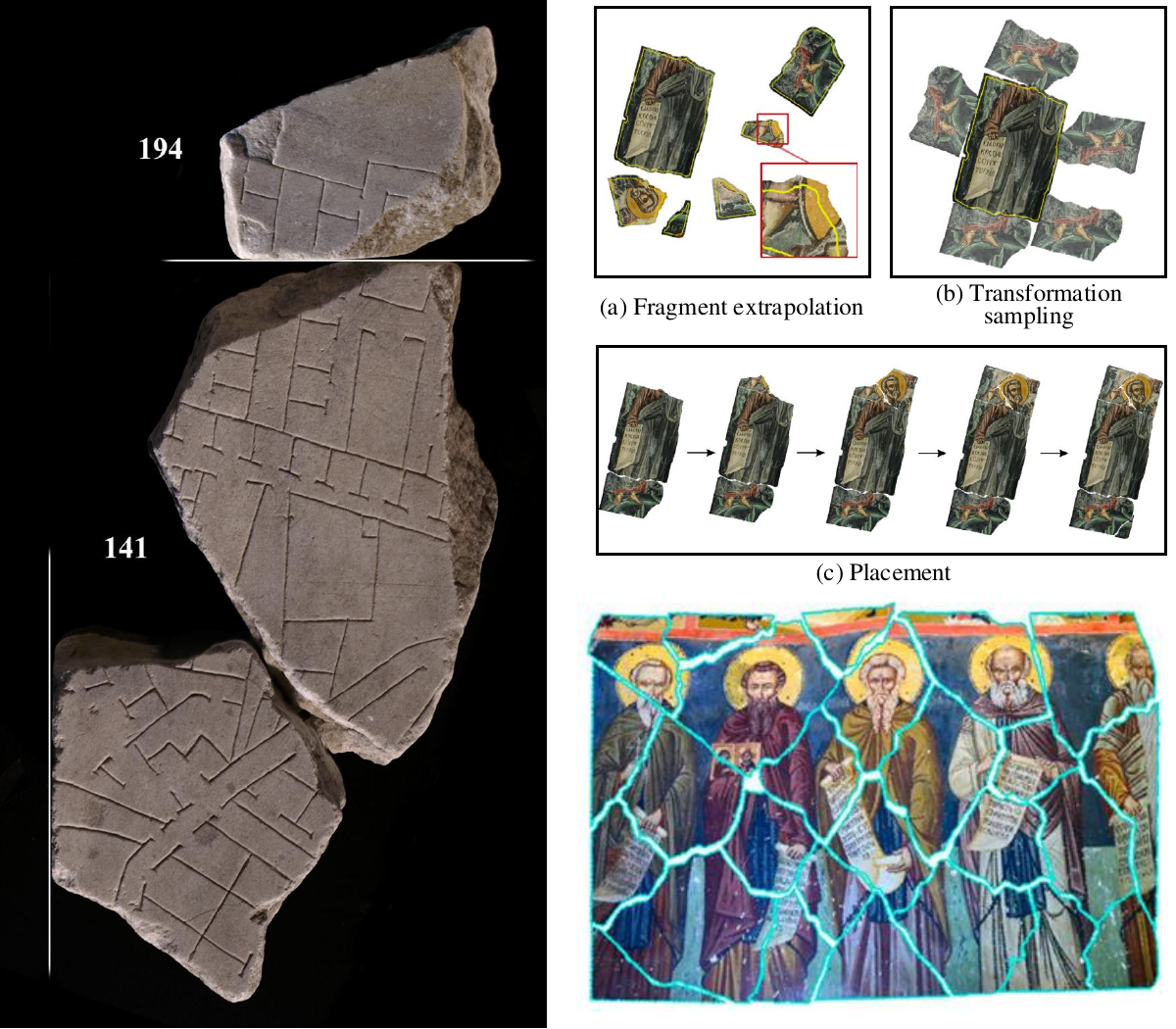}
\caption{Matching features on original surfaces. (Left) Stanford project~\cite{Stanford:2006}. (Right) The idea of matching completed objects with fragment extrapolation~\cite{DBLP:journals/pr/DerechTS21}.}
\label{Figure:Original:Surface:Feature:Matching}
\vspace{-0.1in}
\end{figure}
Most approaches use features extracted from the fracture surfaces. This requires fracture surface segmentation as discussed in Section~\ref{Subsec:Segmentation}. Robustness has been a major concern when extracting features from fracture surfaces.  Therefore, many features introduced in the literature are patch features that are developped to improve robustness. \cite{10.1145/1141911.1141925} use patches defined by sub-level sets of the integral invariant descriptors and summarize geometric descriptors derived from performing principal component analysis of patch shapes.  \cite{DBLP:journals/access/LiGZ20} introduced an approach to smooth the fragments and segment the concave and convex regions of the fracture surfaces for robust matching.  \cite{DBLP:journals/access/JiaHYHCH20} studied multi-scale covariance matrix based point descriptors. \cite{DBLP:journals/ijcv/WinkelbachW08} introduced an algorithm based on cluster-tree tailored for matching fragments with salient fractured surfaces. A key issue in matching fracture surfaces is developing descriptors that can be used to identify geometric patches that can be tucked together~\cite{10.1007/s00371-014-0959-9}. A recent paper~\cite{lu2023jigsaw} also addresses this issue by introducing a primal-dual descriptor in the deep learning approach. 

Many papers focus on extracting boundary curves of fracture surfaces~\cite{10.1007/s00371-014-0959-9,DBLP:journals/access/LiGZ20}. Such features are more robust in the setting where the areas of the fracture surfaces are small, making it difficult to compute and extract geometric descriptors. \cite{PAPAIOANNOU2003401} performs (potentially partial) contour matching of the break boundaries to constrain the pairwise surface registration. \cite{Shin:2012:AAS} analyze and simulate fracture patterns of Theran wall paintings. It builds statistical models of boundary curves of wall painting fragments, which can be applied to match boundary curves. \cite{10.1007/s00371-014-0959-9} detect boundary curves based on curvature computation and apply FFT to compute the descriptors of points on the boundary curves for fragment matching. \cite{Oxholm:2013:Thin:Artifacts} introduced a multi-channel boundary contour representation for matching sub-contours via 2D partial image registration.\cite{DBLP:journals/cg/CohenLE13} perform curvature computation to detect parabolic contours, whose descriptors are determined by moments. \cite{isprs-annals-II-5-393-2014} studied how to detect boundary curves of thin fragments. Fragment matching is done using voting. \cite{DBLP:journals/vc/ZhangCSXZJZB17} introduced a fast algorithm for 2D fragment assembly based on a partial EMD distance. \cite{10.1007/s00371-017-1419-0} introduced a descriptor based on the convex/concave information of a point on the boundary curves of the fractured surface. The similarity between two fractured surfaces is calculated based on the spin images of the point of the feature curve and the distance and normal deviation between the two feature curves to find the matching fractured surface. \cite{alagrami2023reassembling} utilized the detected breaking curves as cues for fracture surface segmentation and registration. \cite{2024:Flake:Surface} introduced a new matching algorithm for stone tool reassembly based on detecting and matching contour points on flake surfaces. Matching is done using a five-point method, which extends Super4PCS. For efficiency concerns, it focuses on matching flake surfaces to a core stone, which is applicable in this specific setting. In \cite{10.1145/3417711}, a set of matching units is detected and described by the 2D Link-Chain Descriptors (LCD) and the 3D Spatial-Distribution
Descriptors (SDD). Second, the pairwise reassembly probability is calculated using LCD and SDD
descriptors; then, collision detection is performed to eliminate the incorrect overlapping pairs. 

Several approaches rely on the features on the original surfaces for matching (see Figure~\ref{Figure:Original:Surface:Feature:Matching}). Early work by~\cite{DBLP:journals/spm/WillisC08} discussed how to use original surface features for geometric matching. Examples include the continuity of texture patterns or matches between curve features extracted from original surfaces. \cite{Oxholm:2013:Thin:Artifacts} introduced a similar approach to enforce the original surface consistency. The former approach can be found in most reassembly pipelines of wall painting fragments, e.g., \cite{Stanford:2006}. \cite{DBLP:journals/pr/DerechTS21} studied how to match fragments that are under erosion and proposed an approach that completes the original surfaces of the fragments and matches the completed surfaces. In this approach, the original non-overlapping surfaces become overlapping, allowing us to detect and match extracted feature points. 

\begin{figure}[b]
% \vspace{-0.1in}
\includegraphics[width=\linewidth]{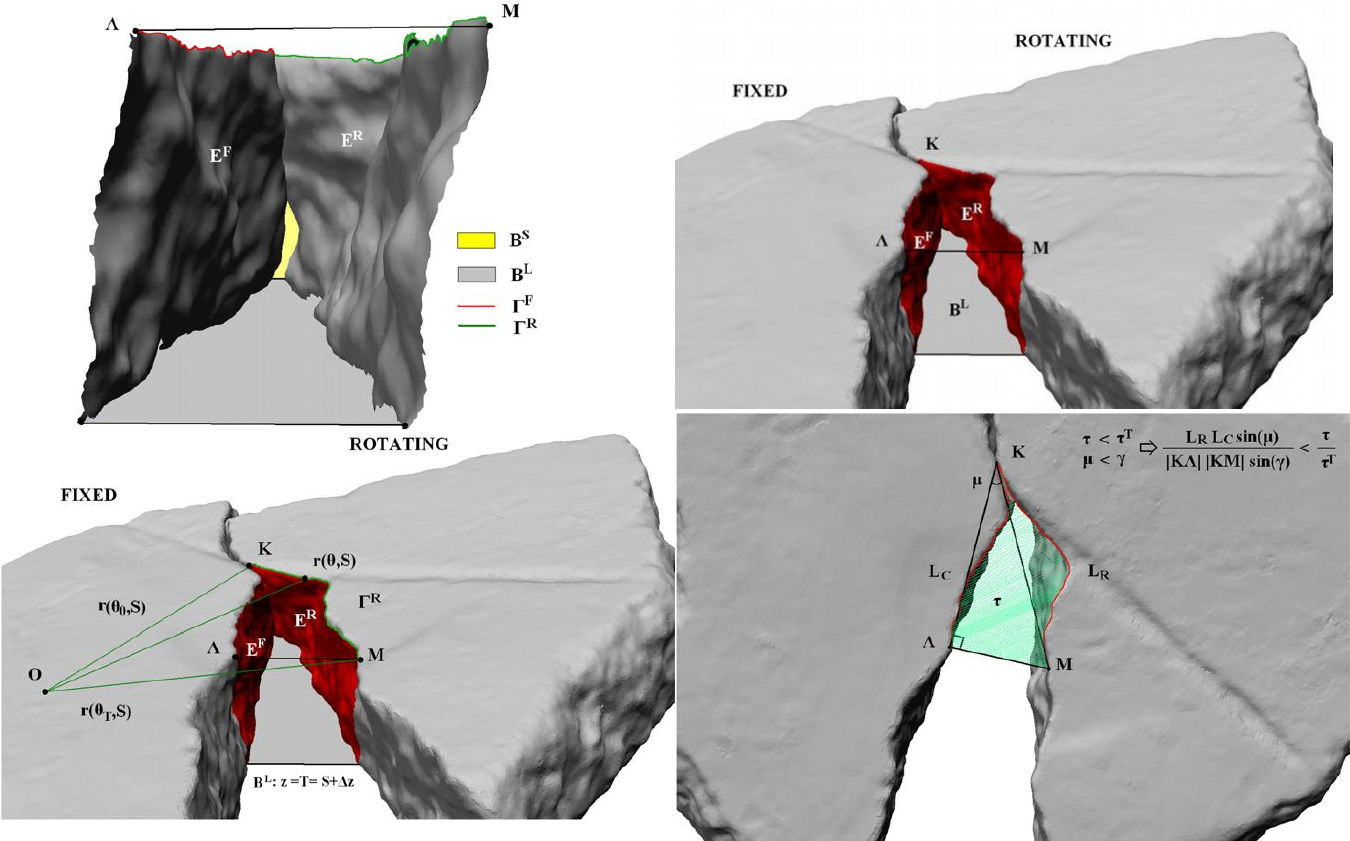}
\caption{Robust scoring functions to evaluate the quality of a match proposed in~\cite{10.1016/j.camwa.2012.08.003}. (Top-left) Criteria 1, gap volume. (Top-right) Criteria 2, contact area. (Bottom) Criteria 3 and 4, bounds for the area and the mean curvature, and the length of contact curves.}
\label{Figure:Robust:Scoring:Function}
\end{figure}

In many cases, the approach of extracting and matching features is not robust when fragments are undergoing serious erosion, which is the case for most real datasets. In \cite{6340022}, the authors find that detecting and matching feature points from fractured surfaces is much more challenging than matching scans. They proposed confidence scores on potential matches. Several approaches define scoring functions over the space of transformations and develop global search strategies for global optimal solutions. \cite{982888} uses simulated annealing to search the space of the source to transformation, while employing hardware-accelerated rasterization to extract surface distances based on depth maps and compute a similarity score using distance gradients. \cite{DBLP:journals/jocch/VidalB14} introduced an approach that focuses on wall-painting fragments and applies a discrete search of relative poses to identify global solutions to reassembly.  \cite{10.1016/j.camwa.2012.08.003} introduced four criteria for fragment matching (see Figure~\ref{Figure:Robust:Scoring:Function}). The first criterion exploits the volume of the gap between two properly placed fragments. The second considers the overlap of the fragments in each possible matching position. Criteria 3 and 4 employ principles from the calculus of variations to obtain bounds for the area and the mean curvature of the contact surfaces and the length of contact curves, which must hold if the two fragments match.

\begin{figure}
% \vspace{2in}
% \vspace{-0.1in}
\includegraphics[width=\linewidth]{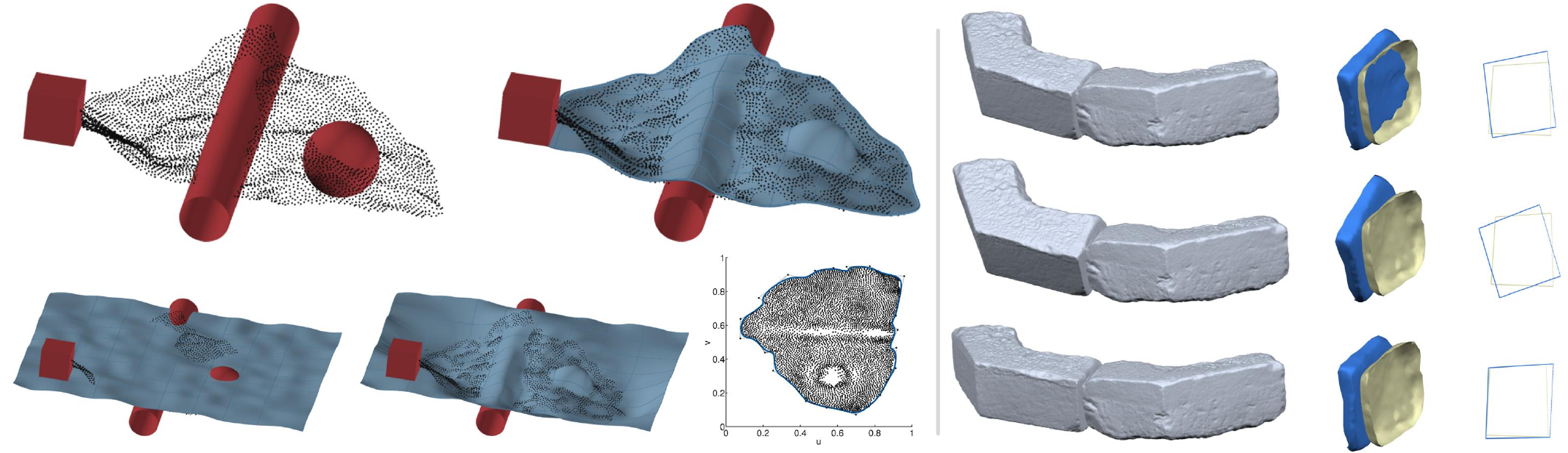}
\caption{Enforcing non penetration via constrained optimization~\cite{10.1145/1141911.1141925,DBLP:journals/cagd/Flory09,DBLP:journals/cad/SchleichW15,DBLP:journals/cad/SchleichW18}.}
\label{Figure:Non:Penetration:Constraint}  
\vspace{-0.1in}
\end{figure}

In addition to geometric matching, another difference between scan matching and fragment matching is that the matches should be free of penetrations. This constraint is easy to integrate for RANSAC and stochastic global optimization approaches~\cite{DBLP:journals/cgf/SchnabelWK07,DBLP:conf/eurographics/MavridisAP15}. An open question is how to enforce this under global optimization approaches such as spectral matching. Regarding local refinement methods, several papers~\cite{10.1145/1141911.1141925,DBLP:journals/cagd/Flory09,DBLP:journals/cad/SchleichW15,DBLP:journals/cad/SchleichW18} studied how to formulate constrained optimization techniques to enforce non-penetration constraints (See Figure~\ref{Figure:Non:Penetration:Constraint} (Left)). \cite{thuswaldner2009digital} introduced a gradient field to optimize the packing of a collection of objects, enforcing the non-penetration constraint (See Figure~\ref{Figure:Non:Penetration:Constraint} (Right)). 

Finally, several systems employ users in the loop and develop semi-automatic approaches for fractured object reassembly. In~\cite{10.5555/2384524.2384531}, user input is used to prune incorrect matches.

\begin{figure*}
% \vspace{2.5in}
\includegraphics[width=\linewidth]{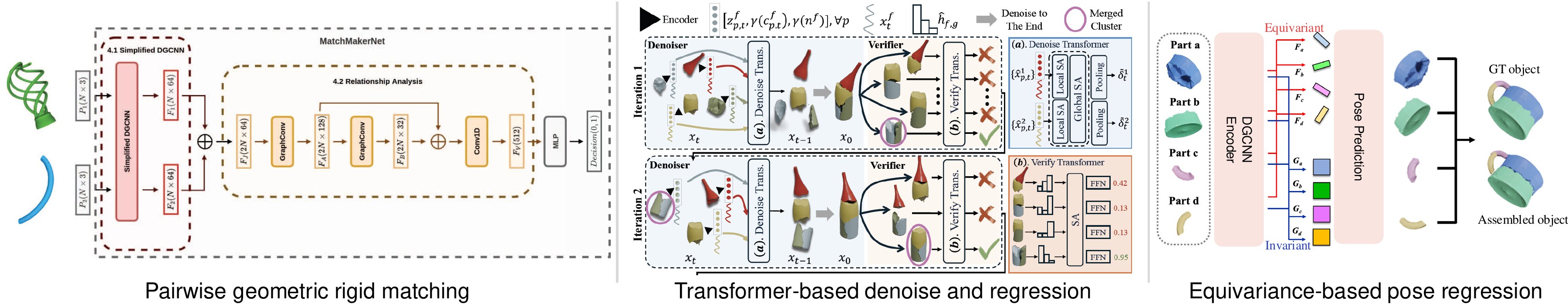}
% \vspace{-0.1in}
\caption{Recent advances on deep fragment matching. (Left) Techniques that extend rigid matching~\cite{DBLP:conf/iccvw/Villegas-Suarez23}. (Middle) Transformer-based techniques~\cite{wang2024puzzlefusionautoagglomerative3dfracture}. (Right) Equivariant techniques~\cite{Wu_2023_ICCV}.}
\label{Figure:DeepLearning:Matching}    
\vspace{-0.1in}
\end{figure*}

\noindent\textbf{Deep fragment matching.} In the deep learning era, the research community has studied developing deep neural networks to solve the multi-piece reassembly problem. Early work includes Deepzzle~\cite{DBLP:journals/tip/PaumardPT20}, which studied image reassembly, a special case of fractured object reassembly. Deepzzle explored methods to handle real-world challenges such as erosion, missing pieces, and the presence of unrelated fragments. Due to the emergence of 3D benchmark datasets for fractured object reassembly, there is increasing interest in developing deep learning approaches to solve the 3D fractured object reassembly. \cite{DBLP:conf/iccvw/Villegas-Suarez23} proposed MatchMakerNet (see Figure~\ref{Figure:DeepLearning:Matching}(Left)), a network architecture designed to automate the pairing of object fragments for reassembling. The approach takes two point clouds as input and its network leverages graph convolution alongside a simplified version of DGCNN for prediction. 
Current approaches have primarily focused on feature learning while incorporating rigidity constraints and domain-specific knowledge remains an important direction for future research. Bridging deep learning methods with practical applications continues to be an active area of investigation.
%First, features and feature matches are usually extracted from fractured regions and their boundaries, in which fractured region segmentation is important. Some other approaches~\cite{DBLP:journals/jocch/PapaioannouSAMG17} leverage continuity of curve and texture features on the original surfaces for matching. These approaches apply to thin-shell fragments and wall painting fragments where fracture regions are relatively small. Second, in the scan matching setting, the features should match under rigid transformations; in contrast, in the fragment matching setting, the features are complementary under rigid transformations. 

%Similar to rigid-scan matching, early fragment matching techniques develop hand-crafted features for matching. Huang et al.~\cite{10.1145/1141911.1141925} compute integral invariant descriptors among fractured surface regions and their boundary curves and apply forward search for rigid scan matching. \qixing{Talk about non-deep learning approaches.}

%\qixing{Talk about deep learning approaches.}
%Dataset (CVPR 2023)~\cite{Lamb-2023-CVPR}

\subsection{Multi-Piece Matching}
\label{Subsec:Multi:Piece:Matching}

%\qixing{Qixing. Bullet points are in.}

\noindent\textbf{Optimization-based based approaches.} The difference between multiple-piece matching and pairwise matching is the cycle-consistency constraint~\cite{DBLP:conf/cvpr/ZachKP10,DBLP:journals/cgf/NguyenBWYG11,DBLP:journals/tog/HuangZGHBG12} among cycles of fragments. Specifically, when composing relative transformations $T_{ij}$ along a cycle of fragments $(i_1, i_2, ..., i_k)$, we should have:
\begin{equation}
T_{i_1i_2}T_{i_2i_3}...T_{i_{k-1}i_k}T_{i_ki_1} = I
\end{equation}
\noindent where $I$ is the identity transformation. If this constraint is violated, then at least one of the relative poses along the cycle must be incorrect. This cycle-consistency constraint is used to prune incorrect pair-wise matches computed between pairs of objects in isolation. 

In the literature on scan matching, early approaches enforce combinatorial optimization to search for consistent matches. A popular way is to optimize a spanning tree among the input graph of pairwise matches~\cite{Huber-2002-8601,DBLP:journals/ivc/HuberH03,10.1145/1141911.1141925}.  The advantage is that it is easy to incorporate the non-penetration constraint among fragments. For example, in the case of scan matching Huber and Herbert~\cite{DBLP:journals/ivc/HuberH03} showed how to incorporate the visibility constraint in multi-scan matching. On the other hand, combinatorial optimization approaches are typically greedy and can easily return sub-optimal solutions. 

Another category of approaches formulates multi-piece matching as Markov-random field (MRF) inference. This is formulated as sampling a discrete set of candidate poses of each object and model pairwise matches as pairwise potentials to optimize the best candidate of each object. This approach was first introduced to solve the Jigsaw puzzle problem in images~\cite{DBLP:conf/cvpr/ChoBAF08,DBLP:journals/pami/ChoAF10}, where the candidate positions of each image patch are given by the grid of image patches of a given image. A relevant approach is \cite{DBLP:conf/bmvc/YuRA16}, which presents a linear programming (LP) relaxation to the MRF formulation on the image-based square patch puzzle problem. This MRF formulation was later applied in MRF-SFM~\cite{DBLP:conf/cvpr/CrandallOSH11,DBLP:journals/pami/CrandallOSH13} to solve the structure-from-motion problem. The nice properties of this MRF formulation are that it possesses better solvers that offer better solutions compared to greedy approaches. Moreover, it is easy to incorporate the non-penetration constraint, i.e., if two fragment penetrates under two candidate poses, then the corresponding pairwise potential is infinite. The limitation of this MRF formulation is that it requires many pose samples to obtain accurate solutions.

Modern multi-piece matching typically formulates low-rank matrix recovery by establishing the equivalence between the cycle-consistency constraint and the fact that the matrix that encodes pairwise maps in the block is low-rank, c.f.,~\cite{DBLP:journals/cgf/HuangG13}. The optimization approaches fall into categories of semidefinite programming relaxation~\cite{SingerA2011Asbe,Wang:2013:IMA,DBLP:journals/cgf/HuangG13,DBLP:journals/tog/HuangWG14,DBLP:conf/icml/ChenGH14,Rosen2019SESync,dellaert2020shonan}, spectral techniques~\cite{Singer:2012:VDM,Kim:2012:ECM,DBLP:journals/tog/HuangZGHBG12,DBLP:conf/nips/PachauriKS13,DBLP:conf/nips/PachauriKSS14,NIPS2016-6128,DBLP:conf/eccv/SunLHH18, arrigoni2016spectral, arrigoni2016camera, bernard2015solution,huang2019learning,DBLP:conf/cvpr/0007H23}, and non-convex optimization~\cite{DBLP:conf/iccv/ChatterjeeG13,zhou2015multi,conf/cvpr/ChoiZK15,DBLP:conf/icra/LeonardosZD17,DBLP:conf/nips/HuangLBH17,DBLP:conf/cvpr/LiSL22,DBLP:conf/iclr/0005HSBH24}. Their advantages are superior empirical performance and theoretical guarantees. However, it is not easy to enforce the penetration-free constraints under them. 

\begin{figure}[t]
% \vspace{3in}
\includegraphics[width=\linewidth]{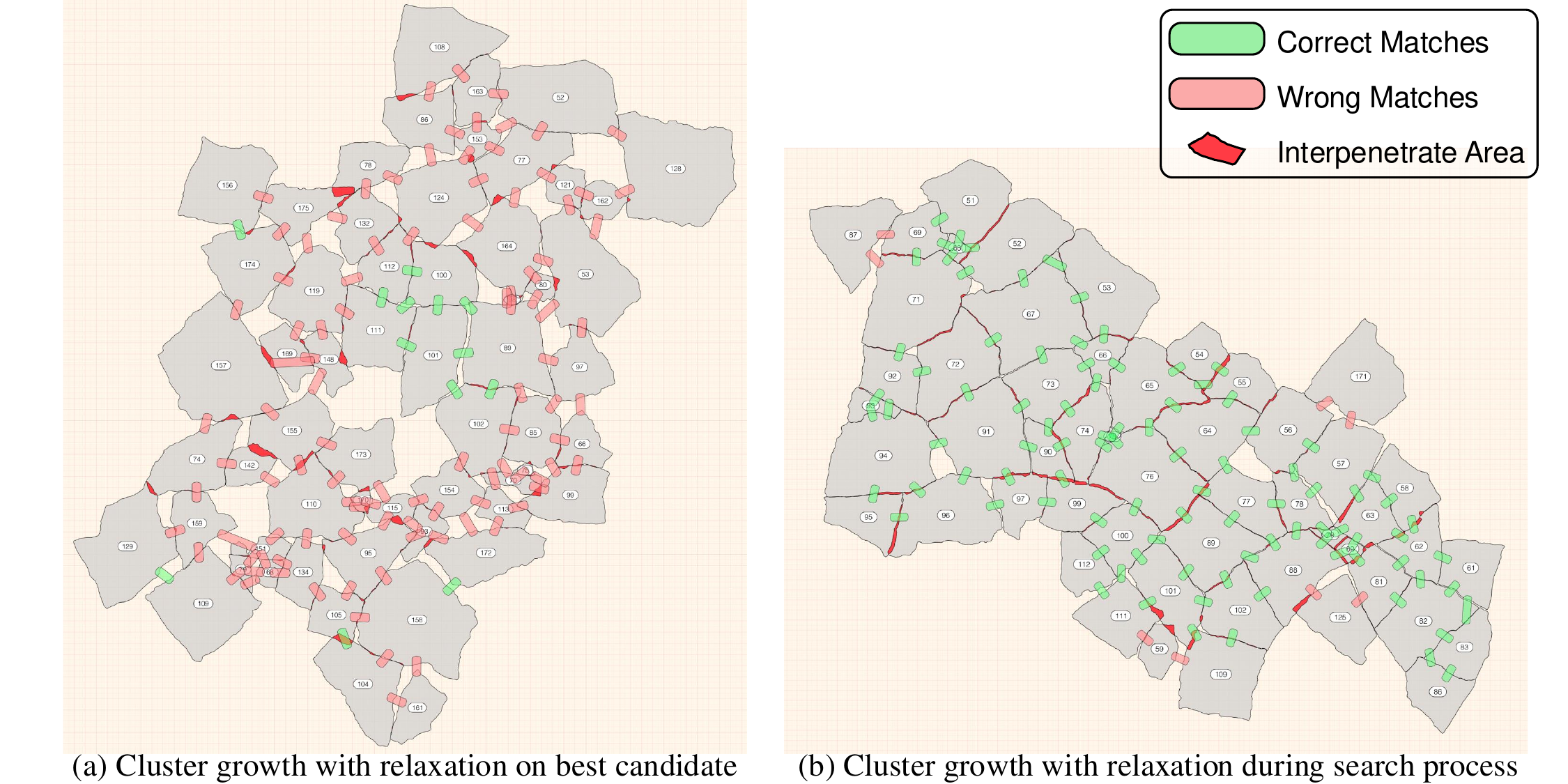}
\caption{Multi-fragment matching~\cite{Castaneda:2011:GCI}, which enforces the consistency of matching along loops.}
\label{Figure:Multi:Piece:Matching}
\vspace{-0.1in}
\end{figure}

Applications of these multi-object matching approaches in the context of multi-fragment reassembly have been limited. Most approaches follow greedy approaches~\cite{10.1145/1141911.1141925,DBLP:conf/dh/AndreadisPM15,wang2024puzzlefusionautoagglomerative3dfracture}. There are a few exceptions. \cite{10.1145/3084547} introduced an unsupervised genetic algorithm to reassemble fragments of wall paintings. The approach evolves a pool of partial reconstructions that grow through recombination and selection over the course of generations. The authors introduced a novel algorithm for combining partial reconstructions that is robust to noise and outliers, and a new selection procedure that balances fitness and diversity in the population. \cite{Oxholm:2013:Thin:Artifacts} detect loop in the pose graph and perform loop closure to rectify pose errors.  \cite{Castaneda:2011:GCI} enforce global Consistency in the Automatic assembly of fragmented artifacts. This approach addresses error accumulation problem the global reassembly of many fragments, among which there are many potential matches with a pair of fragments. 
\cite{10.1145/3417711} introduced a global optimal
reassembly solution is obtained by iterative graph optimization with the constrains of the loop-closures and overlap restrictions. One critical difference between multi-scan matching and multi-fragment matching is that the underlying graph of the fragments is much more sparse than the graph in multi-scan matching, as scans have significant overlap, while fragments do not overlap.

\noindent\textbf{Deep pose regression based approaches.} The emergence of large-scale datasets such as Breaking Bad~\cite{DBLP:conf/nips/SellanCWGJ22} has simulated a wave in developing deep learning approaches that perform pose regression. Many approaches adopt recent advances in 3D feature learning using Transformers. \cite{zhou2023pairingnetlearningbasedpairsearchingmatching} developed Transfomers to find pairs of matching fragments and their relations. Specifically, they first employ a graph-based network to extract the local contour and texture features of the fragments. Then, for the pair-searching task, they adopt a linear transformer-based module to integrate these local features and use contrastive loss to encode the global features of each fragment. For the pair-matching task, they design a weighted fusion module to dynamically fuse extracted local contour and texture features, and formulate a similarity matrix for each pair of fragments to calculate the matching score and infer the adjacent segment of contours. \cite{lee20243dgeometricshapeassembly} introduced Proxy Match Transform (PMT), an approximate high-order feature transform layer that enables reliable correspondences between the mating surfaces of parts while being efficient. Using PMT. they introduced a new framework, dubbed Proxy Match TransformeR (PMTR), for the geometric assembly task. \cite{DBLP:journals/jcst/LiuGJLZY23} introduced a Transformer-based encoder to extract both global and local geometric features, which are then passed to two decoders to estimate the relative pose between two fragments. \cite{wang2024puzzlefusionautoagglomerative3dfracture} introduced a recurrent approach that aligns and merges fragments into larger groups. The key contribution is a diffusion model that simultaneously denoises the 6-DoF alignment parameters of the fragments, and a transformer model that verifies and merges pairwise alignments into larger ones, whose process is repeated iteratively. 
% \cite{xu2024spaformersequential3dassembly} introduced SPAFomer that performs sequential ordering for reassembly. It integrates the ordering encoder, the transformer-based relative encoder powered by positional encoding, and the symmetry encoder that describes geometric relations. Output can be either done in parallel or in an auto-regressive manner. % it's for object assembly...
\cite{DBLP:conf/aaai/ZhangL0DFW24} introduced an approach that identifies critical stones among the input fragments for efficient computation of the reassembly procedure. This is implemented using the attention mechanism in an iterative manner. \cite{DBLP:conf/aaai/CuiYD24} use a hybrid attention module to model and reason complex structural relationships between fragment patches. The model has intra- and inter-attention layers, enabling the capturing of crucial contextual information within fragments and relative structural knowledge across fragments. The model is enhanced with an adjacency aware hierarchical pose estimator.

Researchers have developed equivalent networks~\cite{DBLP:conf/iccv/DengLDPTG21} for point cloud learning to solve fractured object reassembly. \cite{Wu_2023_ICCV} developed equivariant and invariant encoders of each part to compute equivariant and invariant correlation modules to aggregate extracted signals across all parts, which are used to regress the pose of each input part. \cite{wang2024se3biequivarianttransformerspointcloud} introduced an approach, called SE(3)-bi-equivariant transformer (BITR), based on the SE(3)-bi-equivariance prior to the task. Specifically, BITR first extracts features of the inputs using a novel SE(3)×SE(3)-transformer, and then projects the learned feature to group SE(3) as output. BITR can not only handle non-overlapped PCs, but also guarantee robustness against initial positions. \cite{Scarpellini-2024-CVPR} introduced DiffAssemble, which follows Diffusion Probabilistic model formulations. DiffAssemble models a Markov chain in which it injects noise into the position and orientation of the pieces. The correct and random poses at the initial time step and the final time step, respectively. The denoising network is an attention-based GNN that takes as input a graph where each node contains an equivariant feature that describes a particular piece and its position and orientation. Similarly, \cite{fragmentdiff} and~\cite{wang2024puzzlefusionautoagglomerative3dfracture} treated the reassembly task as a denoising task and introduced FragmentDiff and PuzzleFusion++, methods that apply diffusion denoising through the Transformers to predict the pose parameters of each fragment based on their global feature correlations and learned pose priors. While these deep learning approaches show promising results, bridging the gap between synthetic training data and real-world applications remains an important challenge. Real fragments often exhibit erosion, weathering, or missing pieces - conditions that are difficult to fully capture in synthetic training data. Understanding how to better generalize to these practical scenarios is an active area of research.
%\subsection{(to be rearranged)}
%\qixing{Jiacheng}

%A novel approach to reassembling fractured objects is to reformulate the task as a generation problem, which can be effectively addressed using diffusion models \cite{ho2020denoising}. This is precisely the approach taken by \cite{fragmentdiff}. At its core, the method employs a multi-head transformer within the diffusion network to sample poses from the pose distribution, conditioned on features extracted from the fractured parts. These features are obtained using a point encoder based on DGCNN \cite{dgcnn}.

\subsection{Joint Segmentation and Matching}
\label{Subsec:Joint:Seg:Matching}

\begin{figure}[t]
\includegraphics[width=\linewidth]{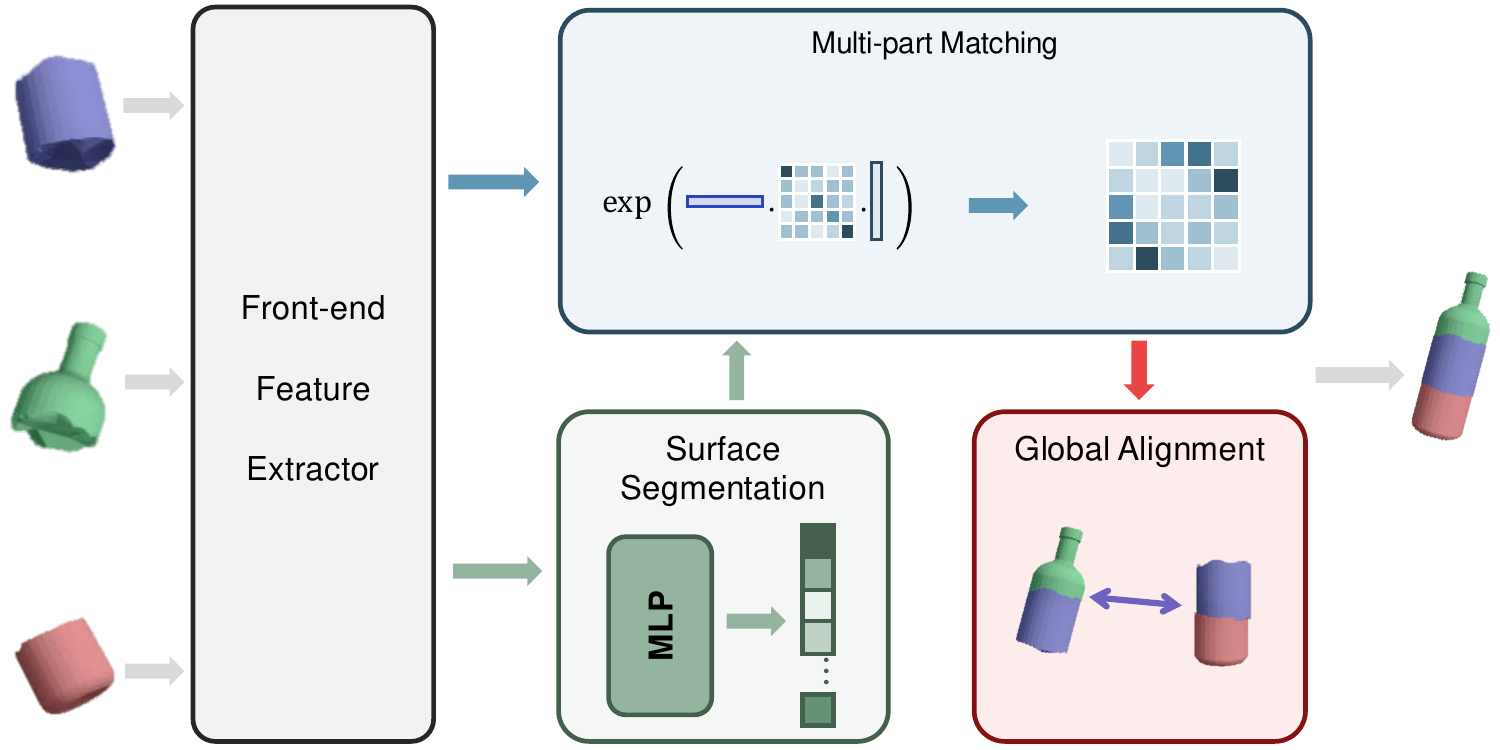}
\vspace{-0.15in}
\caption{Joint segmentation and matching~\cite{lu2023jigsaw}.}
\label{Figure:Joint:Matching:Segmentation} 
\vspace{-0.2in}
\end{figure}

As discussed above, an important problem in fragment analysis is segmenting the fracture surface. Typically, this is done for each fragment in isolation. However, the segmented surfaces should be compatible with those of adjacent fragments. This means that the segmentation problem and the fragment matching problem are correlated and should be solved together. This joint segmentation and matching problem first appeared in the context of joint shape analysis, in which segments of an object provide a compact representation that relates shapes that are structurally relevant but differ in geometric details. \cite{10.1145/2070781.2024159,10.1145/2070781.2024160} introduced joint optimization approaches to optimize the segmentations of individual parts and the correspondences between the segments. \cite{DBLP:conf/iccv/WangHG13,DBLP:conf/cvpr/WangHOG14,DBLP:journals/tog/HuangWG14,DBLP:journals/tog/HuangLWZB19} introduced the functional map framework which allows us to optimize consistent segmentations by optimizing a function sub-space of each shape. 

% In the context of fractured object reassembly, the only approach that explicitly models this objective is JIGSAW~\cite{lu2023jigsaw} (see Figure~\ref{Figure:Joint:Matching:Segmentation}), It extracts point-wise features using transformers, which are used to predict point labels of original or fracture surfaces and perform point-wise matching. Joint segmentation and matching are achieved using shared deep point-wise feature descriptors. Note that such model is often hybrid with procedural methods and utilize RANSAC and off-the-shelf synchronization method to predict transformations. This shows that going purely one way or another, might not by the ideal solution and one should use the right tools at the right place in a synergistic manner. 

In the context of fractured object reassembly, JIGSAW~\cite{lu2023jigsaw} (see Figure~\ref{Figure:Joint:Matching:Segmentation}) pioneered a deep learning approach for joint segmentation and matching. It effectively utilizes transformer architecture to extract point-wise features and proposed primal-dual descriptors for complementary matching. JIGSAW also demonstrates how learning approaches can be effectively combined with classical geometric algorithms. By integrating learned point-wise features with RANSAC and transformation synchronization methods, it achieves robust pose estimation. This hybrid approach leverages both the pattern recognition capabilities of deep learning and the geometric guarantees of classical methods, suggesting that combining complementary techniques may lead to more effective solutions.

%\qixing{Jiaxin}

%\subsection{End-to-End?}
% \end{multicols}
% \begin{multicols}{2}
\section{Template Shape Space}
\label{Section:Template:Analysis}

\begin{figure*}[h]
\includegraphics[width=\linewidth]{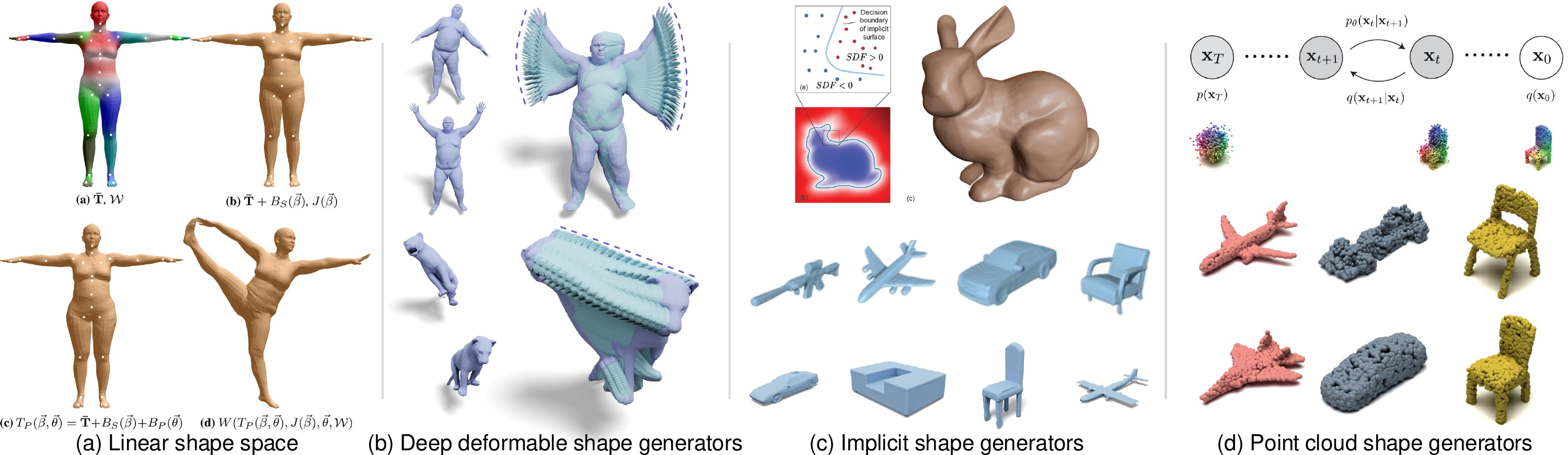}
\vspace{-0.18in}
\caption{(a) Early work on linear shape spaces~\cite{SMPL:2015}. (b) Deep deformable shape generators~\cite{DBLP:journals/tog/YangSCPH23}. (c) Implicit shape generators~\cite{DBLP:conf/cvpr/ParkFSNL19,DBLP:conf/cvpr/ChenZ19,DBLP:journals/tog/ZhangTNW23}. (d) Point cloud shape generators~\cite{Zhou-2021-ICCV}.}
\label{Figure:Template:Shape:Prior:Learning}    
\vspace{-0.1in}
\end{figure*}

%We first discuss approaches that learn general shape spaces in Section~\ref{Subsec:General:Shape:Space}. 
%We then discuss the role of symmetries in encoding template shapes in Section~\ref{Subsec:Symmetries}. 

%completion of an single object~\cite{DBLP:conf/3dor/Sipiran18}.
%Infer from multiple fragments~\cite{DBLP:conf/cvpr/HanCH12}.
%Curvature-based analysis~\cite{DBLP:journals/pr/HanH14}.
%On Estimating the Position of Fragments on Rotational Symmetric Pottery~\cite{DBLP:conf/3dim/SablatnigM99}
%Detection of 3D symmetry axis from fragments of a broken pottery bowl~\cite{DBLP:conf/icassp/YaoS03}.

%More discussions on symmetry detection and symmetry based fractured object reassembly will be discussed in 

%Symmetry refers to inherent geometric properties of the original object that remain invariant under certain transformations, such as rotation, reflection, or translation. which plays an important role in the reassembly of the fractured object. In the context of shape template space, it provides prior knowledge of the structure of the underlying complete object, thus narrowing the search space for matching fragments and helping infer missing pieces and configurations, ensuring a more efficient reassembly.

%\subsection{General Shape Spaces}
%\label{Subsec:General:Shape:Space}

In many cases, we do not have prior knowledge of the exact shape of the underlying complete object. However, we may have prior knowledge about its category, which provides shape priors. Therefore, an important problem is learning a parametric shape space from a collection of example shapes. This problem has been studied extensively in the literature. This section presents relevant approaches that can be applied for fractured object reassembly. We differentiate approaches that focus on organic shapes with the same topology and man-made shapes with varying topology.

Early approaches (see Figure~\ref{Figure:Template:Shape:Prior:Learning}(a)) include Morphable faces~\cite{10.1145/3596711.3596730} and SCAPE~\cite{DBLP:journals/tog/MitraGP06}, which focus on 3D shape spaces with the same topology. Recent examples along this line include SMPL~\cite{SMPL:2015} for humans, SMAL~\cite{Zuffi:CVPR:2017} for animals, MANO~\cite{MANO:SIGGRAPHASIA:2017} of hands, and 3DMM~\cite{nonlinear-3d-face-morphable-model} for faces. These models, trained on larger datasets, demonstrate significantly improved performance in solving inverse problems. They decompose shape variations from pose variations - shape variations are captured through PCA analysis, while pose variations (as in SMAL and SMPL) are encoded using skeletal structures. However, a key limitation of these approaches is that they require training data with consistent mesh topology, meaning that all meshes must share the same vertex structure.

Recent approaches have focused on using deep neural networks to learn generative models, which are superior to linear models. Most approaches~\cite{DBLP:conf/eccv/RanjanBSB18,DBLP:conf/iccv/BouritsasBPZB19,DBLP:conf/eccv/ZhouBP20,DBLP:conf/nips/ZhouWLCYSLS20} use graph neural networks to decode a latent code into a mesh. Many approaches focus on network design. ARAPReg~\cite{DBLP:conf/iccv/HuangHSZJB21} and GeoLatent~\cite{DBLP:journals/tog/YangSCPH23} introduced regularization losses that model shape deformations between adjacent synthetic shapes to improve the generalization of the shape generator. Despite significantly improved performance from linear models, these approaches still require consistent inter-shape correspondences as input for learning.

DeepSDF~\cite{DBLP:conf/cvpr/ParkFSNL19} and follow-up work show that shape generative models can be learned from raw data points without correspondences. SALD~\cite{DBLP:conf/iclr/AtzmonL21} shows that MLP-based implicit surfaces can also encode deformable shapes, but generalization of shapes that undergo large pose changes is very limited. GenCorres~\cite{DBLP:conf/iclr/0005HSBH24} addresses the generalization issue under large pose changes by developing an approach to compute dense correspondences between adjacent implicit surfaces, allowing us to apply ARAPReg to define regularization losses. It leads to significantly improved synthetic shapes under novel poses. 

Significant progress has been made on learning generative models to synchronize man-made shapes. Most approaches that produce high-quality synthetic shapes are under broadly defined implicit shape representations (signed distance functions, volumetric grids, and neural fields), These include early efforts~\cite{10.5555/3157096.3157106,DBLP:conf/cvpr/ChenZ19,DBLP:conf/cvpr/IbingLK21,DBLP:conf/eccv/ChengLLSY22,DBLP:conf/cvpr/YanLMLC022,DBLP:conf/nips/0005NW22} under the generative adversial network (GAN)~\cite{NIPS2014-5ca3e9b1} and variational auto-encoder (VAE)~\cite{DBLP:journals/corr/KingmaW13} paradigms and recent results~\cite{DBLP:conf/siggrapha/HuiLHF22,DBLP:conf/cvpr/ShueCPA0W23,Cheng-2023-CVPR,DBLP:journals/tog/ZhangTNW23,Chou-2023-ICCV,DBLP:conf/cvpr/DongZ0YZDBH24} under the latent diffusion paradigm~\cite{Rombach-2022-CVPR}. However, one limitation of using the implicit representation is that it is very difficult to compute the correspondences between the learned template and the input fragments.

One way to address this issue is to learn generative models under explicit representations such as point clouds and meshes. Many approaches have shown promising results for point cloud generators for man-made shapes~\cite{DBLP:conf/icml/AchlioptasDMG18,luo2021diffusion,Yu-2021-ICCV,Zhou-2021-ICCV,DBLP:journals/corr/abs-2212-08751}. Although it is possible to fit a point cloud generator to shape collections with large geometric and topological variations, there are two limitations that prevent them from serving template models for fractured object reassembly. First, most approaches require the shapes to be consistently oriented, which only applies to a very limited number of categories. Generalization behavior under unaligned training instances is much worse. Second, the correspondences induced from the point cloud generator are not good enough across the synthetic shapes, making it difficult to properly define correspondences between the dynamic template and the input fragments. Recent results in mesh representation, such as deep marching tetrahedra~\cite{DBLP:conf/nips/ShenGYLF21}, MeshGPT~\cite{DBLP:conf/cvpr/SiddiquiAATSRDN24}, and MeshAnything~\cite{chen2024meshanything}, also show impressive synthesis results. However, the two limitations of the point cloud representation remain under the mesh representation. An open question is how to address these two limitations.

%\subsection{Symmetries}
%\label{Subsec:Symmetries}

%\begin{figure}[t]
%\vspace{2in}
%\caption{Symmetric priors.}
%\label{Figure:Symmetric:Priors}
%\end{figure}

%In the context of fractured object reassembly, a popular application of enforcing symmetries is the case of pottery shapes, which have strong rotational symmetries. A pottery shape can be approximated by rotating a curve around an axis. Several approaches have focused on recovering the underlying complete object using information extracted from its fragments~\cite{DBLP:conf/3dim/SablatnigM99,DBLP:conf/icassp/YaoS03,DBLP:conf/cvpr/HanCH12,DBLP:journals/pr/HanH14,DBLP:conf/3dor/Sipiran18}. Many of them are based on curvature analysis to detect the rotation axis and extract the underlying curve. We will discuss symmetry-based fractured object reassembly Section~\ref{Subsec:Symmetry:Based:Template:Reassembly}.

%However, none of the existing approaches have examined the problem of learning a shape space of rotational symmetric shapes for reassembly. Given recent progress on generating shape models discussed in the preceding section, a very interesting question is how to learn a rotational symmetric shape from data. 
% \end{multicols}
% \begin{multicols}{2}

\section{Template-Based Reassembly}
\label{Section:Template:Based}

We first discuss approaches that use (approximate) complete object shapes in Section~\ref{Subsec:Complete:Object:Priors}. We then present approaches that leverage symmetric priors for fractured object reassembly in Section~\ref{Subsec:Symmetry:Based:Template:Reassembly}. Finally, we discuss approaches that utilize learned template shape priors in Section~\ref{Subsec:Learned:Shape:Spaces}.

\subsection{Complete object priors}
\label{Subsec:Complete:Object:Priors}

% key: intro of shape prior
Beyond purely feature-based approaches, using complete objects as templates can significantly boost the reassembly process. This is particularly valuable when working with well-documented objects like skulls, pottery, or architectural elements, where intact reference models often exist. Recent advances in machine learning have also made it possible to learn these geometric priors from collections of similar objects, offering more flexible template-based solutions.

\begin{figure*}[t]
    \includegraphics[width=\linewidth]{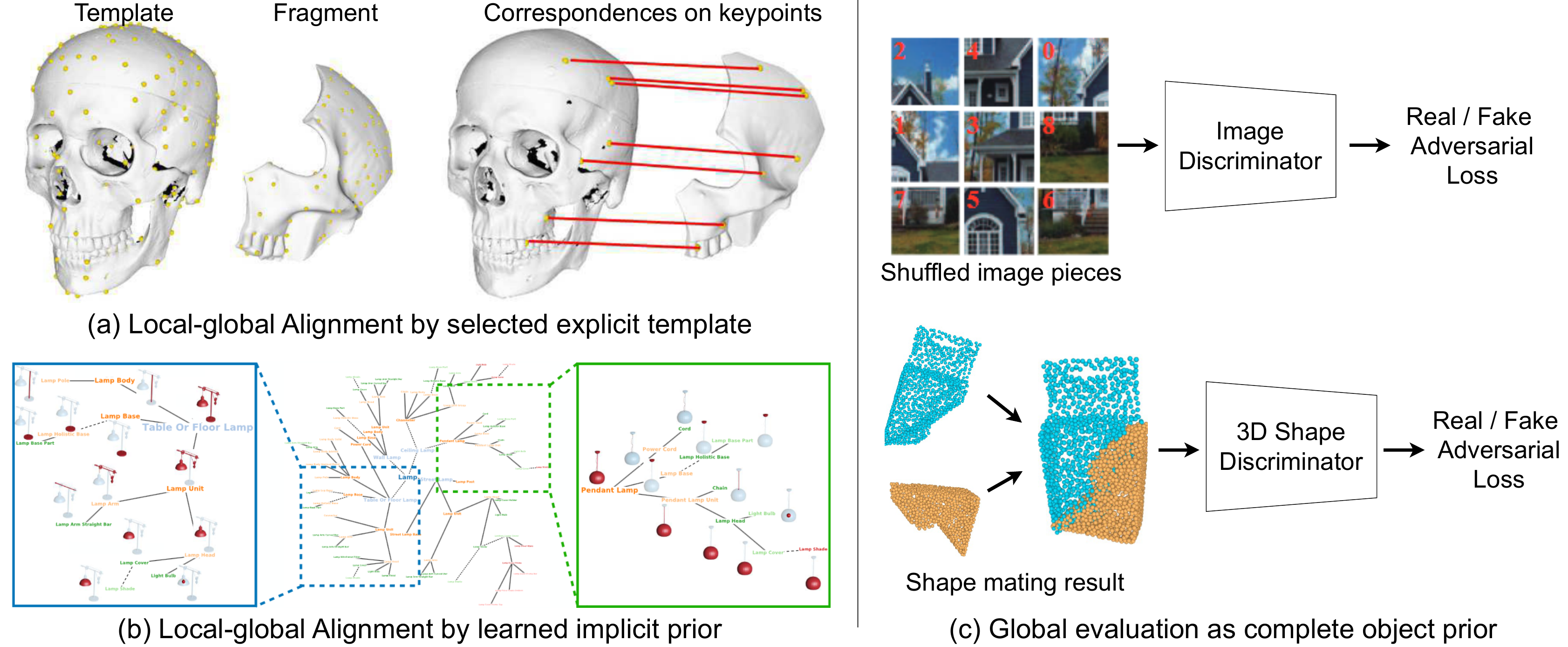}
    \caption{Using prior knowledge from a complete object model to aid reassembly. (a) Local-global alignment with a selected explicit template~\cite{bone:zhang20153d}, such as a skull. (b) Implicit learned prior knowledge from a large-scale 3D object dataset~\cite{Mo_2019_CVPR}. (c) A global evaluation using neural network as complete object prior~\cite{li2021jigsawgan,Chen_2022_CVPR}.}
\label{Figure:Complete:object:priors}
\vspace{-0.1in}
\end{figure*}

\begin{table*}[t]
    \caption{Category of the different complete object priors with their descriptions and limitations.}
    \label{tab:known-prior-object}
    \centering
    \resizebox{\linewidth}{!}{
    \begin{tabular}{c|l|l}
        \toprule
        Categories & Descriptions & Limitations \\
        \midrule
        Local-global Alignment      & Shape prior provides reasonable initial reassembly & Applicable to limited types of reassembly objects\\
        Global Evaluation     & Regularize the quality of the reassembled result & Hard to find a proper function for each reassembly \\
        \bottomrule
    \end{tabular}
    }
\end{table*}

% key: definition and category
This section discusses how a complete object model or global prior can be used to facilitate object reassembly. 
We consider the input fragments as $X=\{x_i\}$ and the global prior as $P$. 
As listed in Table~\ref{tab:known-prior-object}, depending on how $P$ is obtained and used, we categorize relevant methods into two types: methods using \textbf{local-global alignment} and methods using \textbf{global evaluation}.
Local-global alignment methods directly use the global prior $P$ as a guide and solve the matching and alignment between $x_i$ and $P$ to obtain potential compositions of $\{x_i\}$.  
Global evaluation methods do not solve local-global matching, but design a function to evaluate how well a computed reassembly matches the global prior $P$.
Figure~\ref{Figure:Complete:object:priors} illustrates various methods of utilizing complete object priors for reassembly, including explicit template alignment, implicitly learned priors, and network-based global evaluation.

Methods using \textbf{local-global alignment} match fragmented pieces $x_i$ to the global prior $P$ and use this to obtain the initial or rough arrangement of the fragments. 
Their effectiveness depends on three factors:
(1) the design of $P$, 
(2) the modeling/extraction of features from $\{x_i\}$ and $P$ for their matching, and (3) the alignment. 
The global prior $P$ could be a selected/computed \textbf{explicit template} or learned \textbf{implicit parameters}. 
\emph{Classical geometric reassembly methods} often search for a good explicit shape and use it as a template or global prior, then design partial matching schemes to find local-global alignments.  
In contrast, as discussed in Sec.~\ref{Subsec:Learned:Shape:Spaces}, \emph{recent learning-based methods} tend to train neural networks to build implicit shape priors from datasets and then train/optimize the encoding of both local fragments and global template so that their latent codes can infer the matching and alignment of the fragments.

In classical geometric methods, to support effective local-global alignment, the \textbf{explicit shape template} is often selected as an average or typical shape or a best-matched object retrieved from a database.
Using a typical object as the prior shape is common in skull and cultural heritage reassembly tasks.
For example, \cite{yin2011automatic, wei2011skull, bone:zhang20153d} used a standard human skull as $P$ to provide alignment guidance. 
\cite{comes2022digital} used a complete model from the museum's collection as the prior to define the 3D support surface of the fragmented disk.
When a larger collection of complete exemplar objects is available, retrieving a template object with the most similar geometry can provide a more adaptive global prior. 
For example, in the pottery reassembly field, researchers have built many digital databases of pottery.
They support various types of local-global matching for the given fragments, such as drawing sketch~\cite{koutsoudis2010detecting}, a portion of the object geometry~\cite{itskovich2011surface}, and curve descriptors~\cite{DBLP:journals/cg/CohenLE13, smith2014pottery} to retrieve relevant potteries as priors $P$.
In the archaeological and architectural reconstruction of a ruined monument task, to construct the most similar geometry as prior $P$, \cite{thuswaldner2009digital, canciani2013method} conducted a careful study of historical texts that mention the building under consideration.
Given the selected shape prior $P$, as for the second factor, feature extraction computes keypoints and their descriptors on both fragments and the global prior, such as depth and axial symmetry descriptor~\cite{koutsoudis2010detecting}, salient points~\cite{itskovich2011surface}, parabolic contour~\cite{DBLP:journals/cg/CohenLE13}, spin images~\cite{yin2011automatic, wei2011skull}, Signature of Histograms of OrienTations (SHOT) descriptor~\cite{bone:zhang20153d}.
With the extracted keypoints and features, fragment transformations are computed to roughly assemble the input parts to the template.
The researchers developed various alignment methods using keypoint correspondences, such as iterative closed points (ICP)~\cite{yin2011automatic}, geometric distance~\cite{wei2011skull}, local affine-invariant 3D moments~\cite{DBLP:journals/cg/CohenLE13}, Earth mover distance~\cite{itskovich2011surface}, and the RANSAC algorithm~\cite{bone:zhang20153d}.

% key: implicit shape prior
\textbf{Implicit prior} for local-global alignment is learned from an annotated 3D dataset to solve the partial matching from the fragments to the global shape.
The shape prior is encoded in a neural network, which is trained on a 3D segmentation dataset, and the weights are adjusted to learn the semantic relationship between the fragments and the complete objects.
In the inference stage, given a set of parts $X$, trained networks are used as encoders to encode the parts into latent codes, which inherit the implicit shape prior to the possible complete object $P$.
Chaudhuri et al.~\cite{Chaudhuri2011Prob} trained a Bayesian network on a 3D segmentation dataset to encode semantic and geometric relationships among shape components.
Then a probabilistic model is used to compose and align the input parts.
Mo et al.~\cite{Mo_2019_CVPR} proposed a more comprehensive 3D object dataset, named PartNet, a large-scale dataset of 3D objects annotated with hierarchical and instance-level 3D part information.
Following the PartNet dataset, Zhan et al.~\cite{zhan2020generative} and Narayan et al.~\cite{Narayan-2022-WACV} trained an encoder on it to encode the query parts to semantic latent codes, which inherit the shape prior to the complete object.
They framed the 3D part assembly problem as a graph-learning problem.
They took the latent codes as input and used graph neural networks to assemble the parts into complete objects.
They demonstrated their work on 3D chairs, tables, and lamp reassembly tasks.

% key: global evaluation
\textbf{Global evaluation} as the complete object prior $P$ provides a regularization term to constrain the reassembled results. 
It has two variants: hand-crafted functions or learnable discriminators.
Hand-crafted metrics are widely adopted in earlier reassembly tasks.
In the image stitching task, an entropy-based metric is proposed to assess the quality of the stitched panoramic image~\cite{okarma2021entropy}.
In the document reassembly task, Optical Character Recognition (OCR) is used for word or sentence detection~\cite{paixao2018exploring, liang2019reassembling}.
These methods use pair-wise features to suggest candidate alignments of the fragments.
They keep the candidate alignments when the evaluation results are promising, while they discard the alignments when the evaluation results are poor.
More recent approaches leverage large-scale datasets to train neural networks that can evaluate assembly quality. These networks learn to distinguish between correct assemblies and incorrect ones, helping guide the reassembly process toward more plausible configurations.
Li et al.~\cite{li2021jigsawgan} propose a GAN to solve the Jigsaw puzzles.
They designed a network-based discriminator and an adversarial loss to classify the an image is reassembled or real.
The adversarial loss pushes the GAN to assemble images from pieces that look natural.
Then a discriminator was trained as adversarial loss to optimize the mating configuration that should look visually realistic like an object.

\begin{figure}[t]
\includegraphics[width=\linewidth]{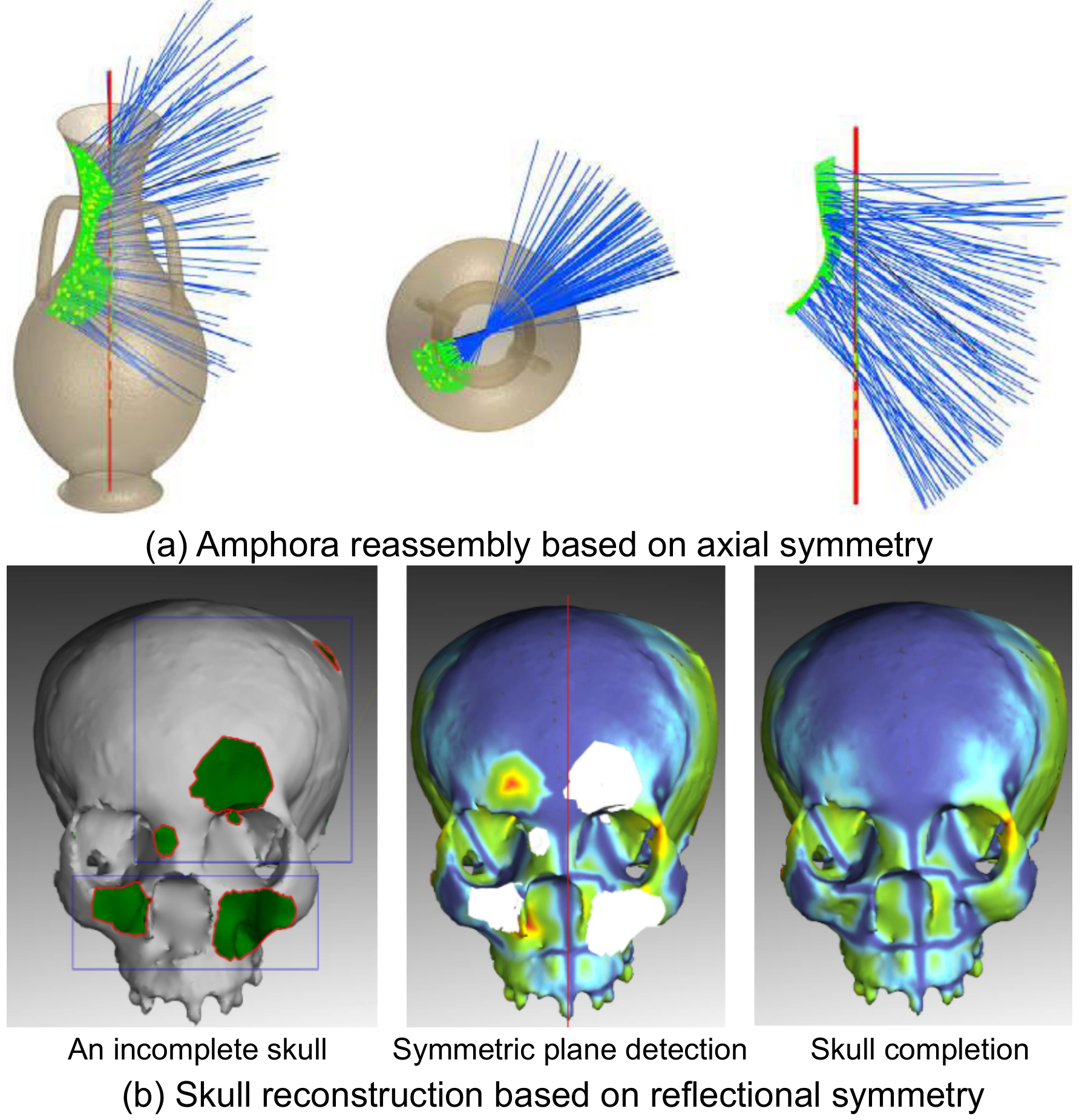}
\vspace{-0.2in}
\caption{Symmetry-based reassembly: identification and alignment using axial symmetry~\cite{NASIRI2022108805} (top) and completion based on reflectional symmetry~\cite{yin2011automatic} (bottom).}
\label{Figure:Symmetry:Based:Reassembly}
\end{figure}

\begin{figure*}
% \vspace{2in}
\includegraphics[width=\linewidth]{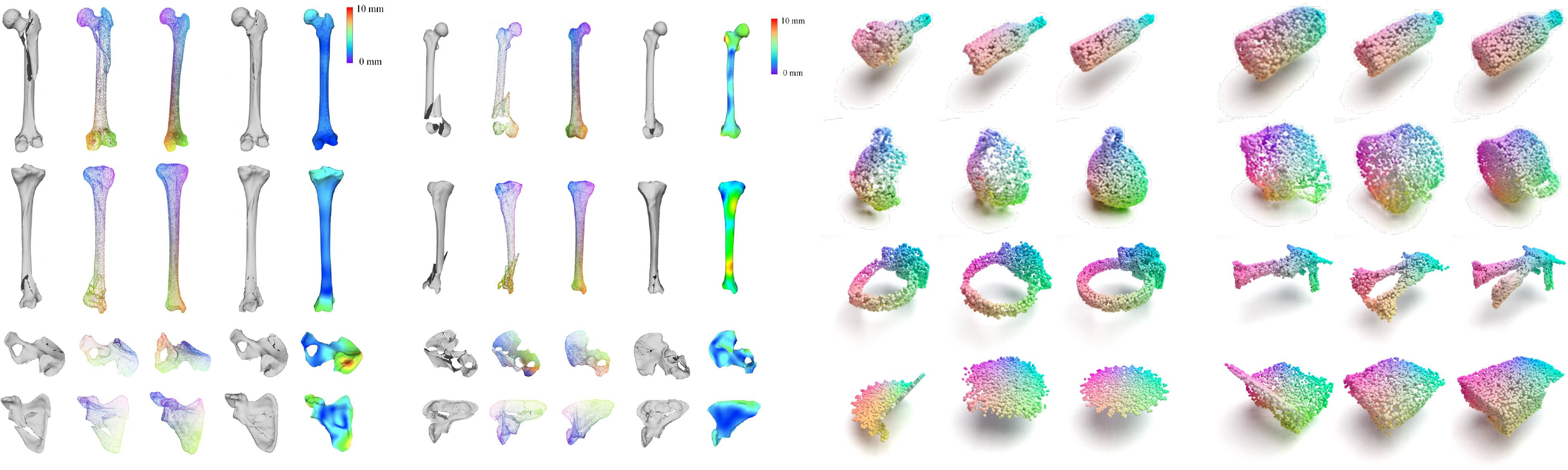}
\vspace{-0.15in}
\caption{(Left) TAssembly~\cite{bone:Deng2023TAssembly}: From left to right – input, correspondence quality, assembly result, and the synthetic template shape. (Right) Jigsaw++~\cite{lu2024jigsawppimaginingcompleteshape}: From left to right – input, Jigsaw++ synthetic result, and the ground truth object.}
\label{Figure:Learned:Shape:Spaces}
\end{figure*}

% key: discussion of limitations
Each approach offers distinct advantages for template-based reassembly. Local-global alignment is particularly effective when working with well-defined object categories that have representative complete objects or comprehensive reference databases. The success of these methods depends heavily on how effectively they can capture relationships between fragments and complete objects. Meanwhile, global evaluation approaches provide flexible guidance for assembly without requiring explicit templates, although the design of effective evaluation metrics remains an active research challenge. Future advances may come from combining these complementary approaches.

% In summary, local-global alignment often works well on a specific category that has a typical complete object or a large enough database as reference.
% The quality of the initial alignment depends on how to encode the part-global relationship into features.
% Due to the lack of well-annotated databases, local-global alignment might be hard to generalize to other object reassembly tasks.
% In contrast, the global evaluation function provides a global shape guidance to regularize the reassembled results. However, finding a proper evaluation function for efficient reassembly is still challenging.

\subsection{Symmetries}
\label{Subsec:Symmetry:Based:Template:Reassembly}

\subsubsection{Symmetry Detection}

Axial symmetry and reflectional symmetry are the two commonly studied symmetry types in geometry analysis and processing, especially in 3D reassembly. For a point cloud $P$, symmetry detection aims to find a transformation $s$ (either axial or reflectional) that minimizes:
\begin{equation}
E(s) = \sum_{p \in P} \min_{q \in P} ||s(p) - q||^2 + \lambda R(s)
\end{equation}
where $s(p)$ is the transformed point under symmetry transform $s$, $q$ is its closest point in $P$, and $R(s)$ is a regularizer that ensures the transformation is a valid symmetry. For axial symmetry, $s$ represents rotation around an axis, while for reflectional symmetry, $s$ represents reflection across a plane.

As shown in Figure~\ref{Figure:Symmetry:Based:Reassembly} (top), axially symmetric, also referred to as rotationally symmetric, indicates that if the complete object's exterior surface is intersected by a plane perpendicular to its axis, the resulting cross section is a circle or closely resembles one.
% how to detect 
As a prominent geometrical property, rotational symmetry can serve as a feature to guide the 3D reassembly of fragmented relics, such as sherds. 
Research on symmetry axis estimation has evolved significantly over time. Early approaches focused on using differential geometric properties, including the sphere of curvature~\cite{cao2002geometric} and normal intersection~\cite{pottmann1999introduction} methods.
Subsequently, enhancements of the estimated rotational axis have been proposed using M-estimators~\cite{halir1999automatic} or Bayesian approaches~\cite{cooper2001assembling,willis2004alignment}. 
These early methods, while foundational, often struggled with real-world challenges such as geometric deformation, missing pieces, and surface noise - common issues when working with fragmented objects. 
\cite{ANGELO201847} suggested a workflow for axis estimation based on minimum thickness direction. This method does not rely on differential geometric properties, making it resilient to local defectiveness, noise, and outliers. 
Recent studies~\cite{hong2019potsac,Hong-2021-ICCV,NASIRI2022108805} have broadly adopted the random sample consensus algorithm (RANSAC) to find good initial estimates.
\cite{hong2019potsac} extended the Cao and Mumford's method~\cite{cao2002geometric}, to consider both the inner and outer surfaces.
This technique introduces PotSAC, a two-stage axis estimator based on a variant of the RANSAC algorithm followed by a robust nonlinear least-square refinement, making it capable of handling noisy surface normals.
\cite{Hong-2021-ICCV} applied beam search to explore multitudes of registration possibilities, addressing false matches between sherds inflicted by indistinctive sharp fracture surfaces.
\cite{NASIRI2022108805} showed that the identification of the symmetry axis from surface normal lines may yield two meaningful local minima. A multiple solution RANSAC algorithm is proposed for initial estimates, followed by a coordinate descent algorithm to refine these estimates. The authors showed that their method ensures convergence, achieves a faster convergence rate, and provides state-of-the-art performance in axis estimation.

As illustrated in Figure \ref{Figure:Symmetry:Based:Reassembly} (bottom), reflection symmetry refers to the property that an object does not change upon undergoing reflection.
The human skull typically exhibits reflection symmetry along the mid-sagittal plane~\cite{li2011symmetry}. 
This kind of symmetry is one of the main geometric priors used in skull reassembly and completion~\cite{wei2011skull,yin2011automatic}.

\subsubsection{Symmetry-guided reassembly}

% how to guide reassembly
With an estimated symmetry axis, the consistency of the symmetry axis and the axis profile curve can be viewed as a constraint or optimization objective during reassembling. \cite{Son_2013_CVPR,9412372} implemented a global constraint that all fragments share the same axis, greatly reducing computational costs and improving noise resistance. This method reduces the search space from 6 DOF (for two fragments) to 1 DOF (relative position of two pieces along the axis of symmetry). \cite{DBLP:conf/dh/AndreadisPM15} exploits the results of an ongoing reassembly to predict the shape of the resulting object using symmetry-based expansion and then performs alignment of disjoint parts with the predicted shape to also assemble non-contacting parts.
\cite{Hong-2021-ICCV} relaxed the above constraint by encoding the information about the axis as a descriptor into each break line for efficient reassembly.

\subsection{Learned Shape Spaces}
\label{Subsec:Learned:Shape:Spaces}

Compared to approaches that rely on similar shapes or symmetric templates to guide fractured object reassembly, few methods have explored the use of a generic template shape space learned from the data to enhance this process. One notable exception is TAssembly~\cite{bone:Deng2023TAssembly}, which learns a linear shape space specifically for bones. It formulates fractured object reassembly as an optimization problem, adjusting fragment poses and the underlying complete shape in the learned shape space to minimize the distance between the fragments and the dynamic template. However, this method is limited to bone shape spaces, where consistent mesh parameterization makes it relatively straightforward to learn descriptors for predicting dense correspondences. Another promising approach is Jigsaw++~\cite{lu2024jigsawppimaginingcompleteshape}, which employs a point cloud generation model with a rectified flow-based retargeting algorithm to infer the complete shape from a partially assembled input. Although this approach shows significant potential for reconstructing daily objects, scaling it to handle more diverse or even open-vocabulary input shapes requires larger models and more data. 
Significant challenges remain in developing effective shape priors for objects with diverse geometries and topologies. Additionally, current methods often lack mechanisms for establishing correspondences between reconstructed shapes and input fragments, making it difficult to enforce rigidity constraints between fragments.
% \end{multicols}
% \begin{multicols}{2}
\section{Applications}
\label{Section:Applications}

We discuss four main applications areas for fractured object reassembly (Section~\ref{Subsec:Bone:Reduction} to Section~\ref{Subsec:Scientific:Applications}) and the relevant application in shape modeling via part assembly.

\subsection{Bone Reduction}
\label{Subsec:Bone:Reduction}

\begin{figure}[t]
\includegraphics[width=1.0\columnwidth]{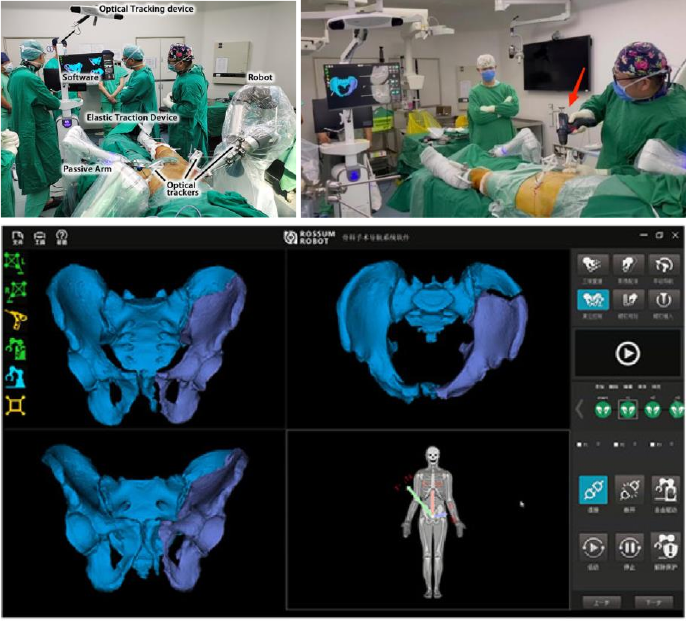}
\vspace{-0.1in}
\caption{The surgical robot designed for bone reduction is detailed in~\cite{bone:zhao2023intelligent,bone:ge2022robot}. At the top, the figure illustrates the components of the surgical robot during a bone reduction procedure within a surgical setting. At the bottom, the figure presents the surgical planning system associated with the robot.}
\label{Figure:Application:bonereduction:surgicalrobots}
\end{figure}

3D bone reduction is essential for preoperative surgical planning~\cite{bone:liu2024end,bone:Yibulayimu23MICCAI,bone:LiuB14} and intraoperative navigation~\cite{bone:mia/Han21,bone:zeng2024bidirectional}, including robotic surgery~\cite{bone:lee2022roboticsytem,bone:schuijt2021robot}(see Figure~\ref{Figure:Application:bonereduction:surgicalrobots}). The objective of bone reduction is to restore the anatomical alignment of fractured bones. Bone fragments have two unique characteristics that distinguish them from other fractured objects: First, human bones have strong shape priors due to their anatomical nature, including bilateral symmetry which allows the use of contralateral bones as templates (see Section~\ref{Section:Template:Based}). Second, bones have complex internal structures including cortical bone, cancellous tissue, and cartilage. When fractures occur in cancellous regions, especially in metaphyseal areas of long bones, the irregular fracture surfaces can make precise alignment challenging~\cite{bone:Deng2023TAssembly,bone:Deng2023Synergistically}. 

\textbf{Skull reconstruction}, as a specialized case in bone reduction, presents unique characteristics and challenges. Unlike long bones, skull fragments are typically modeled as a collection of surfaces, where boundary loops serve as key features for matching~\cite{yin2011automatic,wei2011skull}. Feature descriptors computed on these boundary curves enable reliable fragment matching. The reconstruction process leverages the skull's natural bilateral symmetry~\cite{li2011symmetry}, particularly useful when fragments are missing from one side. For cases involving multiple feature curves on a single fragment, the matching problem can be formulated as the Largest Common Point-set (LCP) problem~\cite{bone:zhang20153d}. Statistical shape models trained on skull datasets provide additional guidance for cases with severe damage~\cite{bone:Deng2023TAssembly}. The precision requirements are particularly stringent due to the skull's critical role in both surgical planning and forensic analysis. 

The bone reduction workflow typically consists of four steps: CT bone segmentation, template matching, pairwise fragment matching, and multi-fragment matching.

\begin{figure}[t]
\centering
\includegraphics[width=0.8\columnwidth]{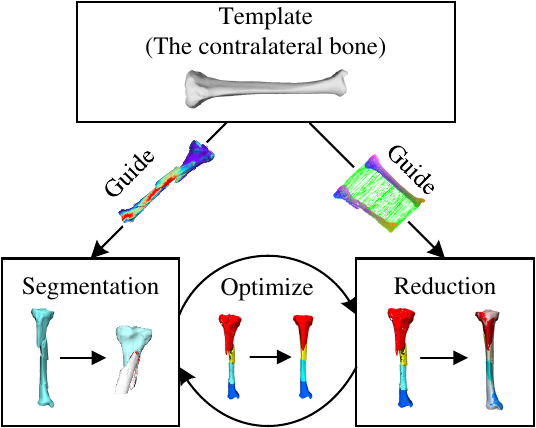}
\caption{Synergistically segment and reassemble the fragments by utilizing the contralateral bone as a template~\cite{bone:Deng2023Synergistically}.}
\label{fig:bone:segred}
\vspace{-0.1in}
\end{figure}

\textbf{CT bone segmentation} endeavors to separate fractured bones from surrounding healthy tissue and divides individual fragments in CT images. A key challenge is distinguishing between fracture zones and intact zones. This can be done using automatic methods like watershed~\cite{bone:liu2024end}, probabilistic watershed~\cite{bone:Liup2021CMIG}, template-matching~\cite{bone:Deng2023Synergistically}, or semi-automatic approaches like region growing~\cite{bone:VlachopoulosSGF18(scale_space),bone:paulano20143d-segregiongrow} and continuous max-flow~\cite{bone:mia/Han21}. Recent deep learning approaches include supervised contrastive learning~\cite{bone:zeng2023two}, synergistic segmentation with contralateral bone templates~\cite{bone:Deng2023Synergistically} (see Figure~\ref{fig:bone:segred}), and dual-stream learning~\cite{bone:zeng2024fragment-seg}. The outer surface of the intact zone~\cite{bone:Liup2021CMIG,bone:Deng2023Synergistically} serves as a template and can be extracted using the MarchingCube algorithm~\cite{lorensen1998marching}.

\textbf{Template matching} provides initial coarse alignment between bone fragments and a template. Most commonly, the healthy contralateral bone serves as a template~\cite{bone:liu2024end,bone:Deng2023Synergistically,bone:okada2008computer,bone:Liup2021CMIG,bone:ead2020virtual,bone:hu2013femur}, though this approach has limitations due to anatomical variations~\cite{bone:vlachopoulos2016computer} and may not always be available. As an alternative, parametric statistical shape models (SSM)~\cite{bone:Deng2023TAssembly,bone:Yibulayimu23MICCAI,bone:mia/Han21} can be used. For complex cases involving joint dislocations, statistical pose models (SPM)~\cite{bone:han2020multi} have been developed. A notable advancement by~\cite{bone:mia/Han21} combines both SSM and SPM in an iterative procedure that alternates between fragment alignment, SSM adaptation, and SPM adaptation.

\textbf{Pair-wise matching} aligns two fragments by matching their fracture boundaries, involving fracture boundary identification followed by boundary matching. Early pioneering work~\cite{bone:willis20073d} introduced a computational framework combining geometric analysis with medical imaging for pairwise bone fragment registration. After template matching brings fragments into rough alignment, fracture boundaries can be identified using Euclidean distance~\cite{bone:liu2024end}. For fractures in the cortical bones, pronounced curvature helps to identify boundaries~\cite{bone:okada2008computer,bone:furnstahl2012computer,bone:kronman2013automatic,bone:buschbaum2015computer,bone:Liup2021CMIG,bone:luque2021complex}. However, fractures in trabecular bone often appear planar, requiring manual annotation~\cite{bone:liao20203clustertree,bone:okada2008computer}. For cortical bone fractures, matching techniques similar to those used in the restoration of cultural artifacts~\cite{DBLP:journals/ijcv/WinkelbachW08,bone:VlachopoulosSGF18(scale_space),bone:buschbaum2015computer} are employed, using criteria like tangential contact points and opposing surface normals. ICP is also commonly used~\cite{bone:willis20073d,bone:okada2008computer,bone:kronman2013automatic,bone:luque2021complex,bone:liu2024end}. For trabecular bone fractures or cases with missing fragments, template matching becomes essential. 

\textbf{Multi-fragment matching} optimizes the poses of multiple fragments globally, using poses from template and pair-wise matching as input. The goal is to maximize region matching scores~\cite{bone:zhang20153d} through either exhaustive~\cite{bone:VlachopoulosSGF18(scale_space)} or greedy search~\cite{bone:pulli1999multiview}. The greedy search method incorporates heuristic rules for matching and nex-bone selection, where various criteria, including template and pair-wise matching scores~\cite{bone:zhang20153d}, fragment-to-set distance~\cite{bone:furnstahl2012computer}, fracture surface area~\cite{bone:luque2021complex}, and fragment size~\cite{bone:liu2024end}. To avoid local minima, beam search can select multiple fragments simultaneously~\cite{bone:zhang20153d}. Error accumulation is addressed through the ideal mates concept~\cite{bone:furnstahl2012computer,bone:Liup2021CMIG,bone:pulli1999multiview} or graph optimization~\cite{bone:zhang20153d}. Similar to stone fragment matching, bone reduction approaches often incorporate domain-specific knowledge.

\textbf{Deep learning models} were trained to segment the cortical surface of bone fragments~\cite{bone:Deng2023Synergistically} and fractured bones~\cite{bone:liu2023pelvicSeg}, identify and localize bone fractures~\cite{bone:zeng2023two}, and compute bone features~\cite{bone:Deng2023TAssembly,bone:Deng2023Synergistically} for template matching. A notable advancement \cite{bone:zeng2024bidirectional} introduced a bidirectional framework combining fracture simulation and structural restoration, where the simulation considers patient-specific anatomy and bone properties to generate training data. This approach uses both simulated scenarios and real fracture data, where contralateral healthy sides provide ground truth, to train a network that reconstructs pre-fracture states for template matching.

% 3D
\subsection{Fractured Cultural Heritage Reassembly}
\label{Subsec:Fractures:cultural:heritage}

Cultural heritage restoration represents a primary application domain for computational reassembly techniques. Reconstructing damaged artifacts preserves historical objects and provides insights into past civilizations and artistic practices. We examine four categories of heritage objects: thin-shell artifacts like pottery, stone fragments from sculptures and buildings, planar objects such as maps and inscribed tablets, and larger-scale archaeological site reconstructions. Each category presents distinct geometric challenges - from pottery's axial symmetry to stone's complex fracture patterns - requiring specialized computational approaches that build upon core reassembly techniques discussed earlier.

\begin{figure}
\includegraphics[width=1.0\columnwidth]{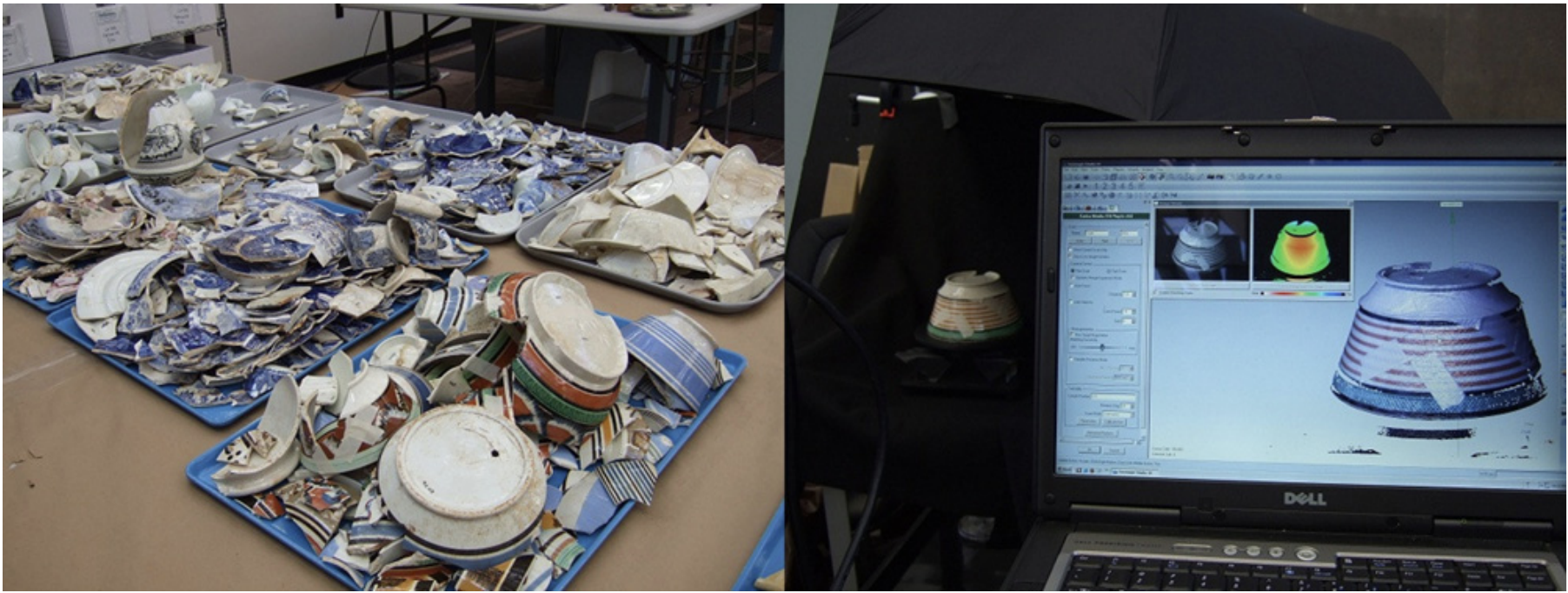}
\caption{Photo from~\cite{DBLP:journals/cg/CohenLE13} that shows a project that reconstruct thin-shell objects from its fragments. (Left) fragments. (Right) A digitized fragment.}
\label{Figure:Thin:Shell:App}    
\vspace{-0.1in}
\end{figure}

\noindent\textbf{Thin-shell objects.}
The reassembly of thin-shell archaeological artifacts, particularly pottery~\cite{DBLP:journals/cg/CohenLE13,10.1145/3569091} and ceramics~\cite{KARASIK20081148,Hong-2021-ICCV}, presents unique challenges due to their distinctive geometric properties. These objects are characterized by surfaces where thickness is negligible compared to other dimensions, requiring specialized computational approaches. Early work by Cohen et al.~\cite{DBLP:journals/cg/CohenLE13} established a foundational pipeline using digitized fragments from 3D scans, computing parabolic contours on surfaces, and employing affine-invariant 3D moments for contour matching and alignment. This approach demonstrated significant speed improvements over manual methods. For ceramics specifically, researchers have leveraged their inherent axial symmetry properties. Karasik and Smilansky~\cite{KARASIK20081148} developed efficient scanning protocols specifically for ceramic fragments, while Hong et al.~\cite{Hong-2021-ICCV} introduced an incremental 3D reassembly method that exploits axial symmetry for more robust matching.

Subsequent research enhanced the reconstruction process through various specialized techniques. \cite{di2018automatic} developed a recognition system that segments fragments into distinct regions (rim, walls, and base), designing region-specific descriptors to guide the matching process. \cite{fu2020automatic} introduced a curve-based matching approach using curvature and torsion descriptors combined with derivative dynamic time warping (DDTW), incorporating voting mechanisms and ICP refinement. A recent breakthrough~\cite{wang2023batch} addressed a critical challenge in thin-shell reassembly - the lack of distinctive geometric features inside regions. Their method combines contour-based batch matching with boundary-aware ICP registration, achieving remarkable efficiency by processing over 700 fragments within 10 hours while maintaining high accuracy. This progression of research demonstrates the evolution from basic geometric matching to sophisticated systems that can handle large-scale reconstruction tasks.

\begin{figure}[b]
\vspace{-0.1in}
\centering
\includegraphics[width=\columnwidth]{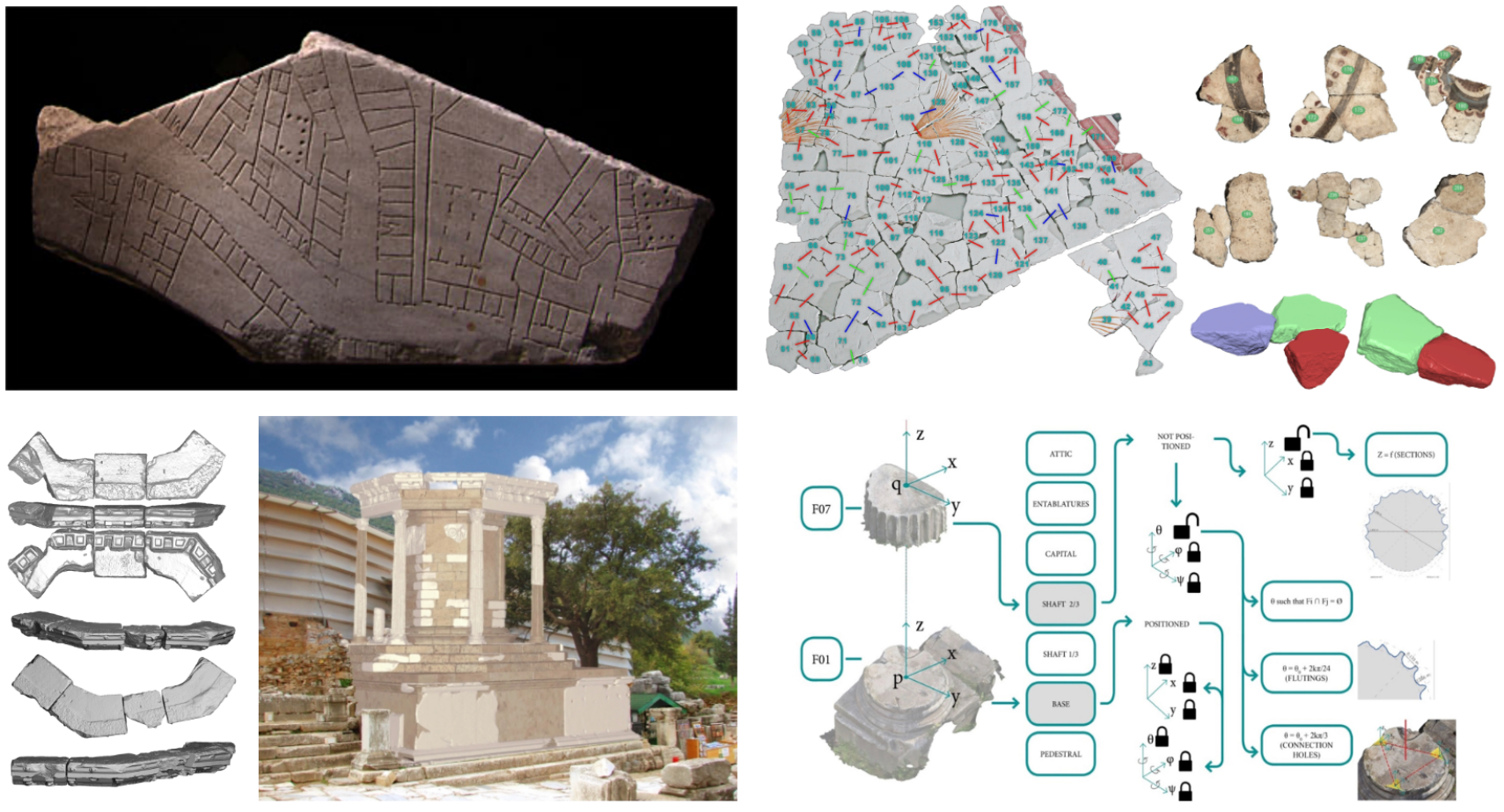}
\caption{The top row shows projects of restoring planar fragments. The bottom row shows projects of restoring building fragments. (Top-left). Stanford's Forma Urbis Romae project~\cite{Stanford:2006:2}. (Top-right) Princeton's . Theran wall paintings project~\cite{10.1145/1360612.1360683}. (Bottom-left). Restoration of the Octagon monument in Ephesos~\cite{thuswaldner2009digital}. (Bottom-right) Reconstruction of the Arch of Titus at the Circus Maximus in Rome~\cite{canciani2013method}. }
\label{Fig:Wall:Building:Fragments}
% \vspace{-0.1in}
\end{figure}

\noindent\textbf{Planar heritage.} 
Planar heritage objects represent a significant category in computational reassembly, encompassing artifacts such as ancient maps, floor plans, frescoes, and inscribed tablets. A landmark project in this domain is Stanford's Forma Urbis Romae project~\cite{Stanford:2006:2} (Figure~\ref{Fig:Wall:Building:Fragments} (Top-Left)), which aimed to reconstruct an ancient marble map of Rome. Despite scanning 1186 fragments (approximately 20\% of the estimated total), the project faced significant challenges due to missing pieces. Their methodology combined multiple matching approaches: boundary incision matching using surface textural features, location prediction using map features, and multivariate clustering that considered characteristics like marble veining direction and fragment thickness~\cite{Stanford:2006}.

The Theran wall paintings project ~\cite{Karianakis2013AnIS,10.1145/1882261.1866207,10.1145/1360612.1360683,10.1145/2037820.2037824,Castaneda:2011:GCI,10.1145/2362402.2362404,10.1145/3084547} represents another significant body of work in planar heritage reconstruction, including a series of work at Princeton University on reconstruction~\cite{10.1145/1360612.1360683}, matching~\cite{10.1145/1882261.1866207,10.1145/2037820.2037824,Castaneda:2011:GCI}, and analysis of fragment boundary patterns for fragment pattern simulation~\cite{10.1145/2362402.2362404}. Moreover, \cite{10.1145/3084547} presents a generic algorithm for reassembling wall paintings. One major difference between wall fragments and fragments of bones and thin-shells is that the number of wall fragments is much larger. In other words, it is important to develop computational solutions to fractured object reassembly. Another consequence of having a large number of fragments is that the global consistency constraint~\cite{Castaneda:2011:GCI,10.1145/3084547} becomes important for multi-fragment matching.

For inscribed tablets, researchers have addressed the challenge of preserving both geometric and textual information. \cite{mara2010gigamesh} developed methods for cuneiform tablet reconstruction considering both wedge impressions and surface geometry, while~\cite{collins2014tablets} introduced a system combining geometric and semantic analysis. More recent work~\cite{Rothacker2015tablets} has used machine learning to analyze cuneiform characters for matching purposes.

Recent advances in this field include RePAIR dataset~\cite{tsesmelis2024reassembling}, which provides both 2D and 3D puzzles from collapsed frescoes, incorporating real-world challenges such as missing pieces and irregular fragment shapes. Related work on flat document restoration, such as papyrus reconstruction~\cite{10.1145/3460961}, has leveraged modern machine learning techniques, combining convolutional neural networks with traditional classifiers to address the unique challenges of document reassembly.

\noindent\textbf{Stone fragments.} 
The reassembly of stone fragments encompasses both individual artifacts like sculptures and large-scale architectural remains, presenting unique challenges due to material properties, scale, and frequent missing pieces. Early work~\cite{10.1145/1141911.1141925} introduced geometric matching techniques specifically designed for fractured objects, combining patch-based features with global optimization to handle stone fragments effectively. For sculptural fragments, \cite{DBLP:journals/spm/WillisC08} developed methods based on original surface features, while \cite{10.1016/j.camwa.2012.08.003} introduced robust scoring functions considering gap volume and contact surface properties. Recent work~\cite{2024:Flake:Surface} proposed specialized methods for stone tool reassembly using contour points and flake surfaces.

In architectural reconstruction, the challenges scale significantly with fragment size and quantity. The Octagon monument project~\cite{thuswaldner2009digital}, employing techniques from~\cite{10.1145/1141911.1141925} for pairwise matching and multi-piece assembly, demonstrated semi-automatic restoration using only 20\% of original fragments. The Arch of Titus at the Circus Maximus in Rome reconstruction~\cite{canciani2013method} leveraged architectural symmetry to handle extensive missing pieces. Similar approaches have been applied to other significant sites, including the Maritime Theatre~\cite{Maritime:Theatre:2019} and Macedonian Tombs~\cite{Tomb:2020}. For specific architectural elements, such as column drums, \cite{10.1145/2901297} developed specialized pattern matching techniques. Recent advances \cite{DBLP:journals/pr/DerechTS21} address eroded fragments by developing completion methods before matching. A key trend across both domains is the integration of computational methods with expert knowledge, particularly crucial when dealing with weathered surfaces or significant missing portions.

%\cite{10.1145/2901297} added to the last paragraph, please check.

\begin{figure}[t]
\centering
\includegraphics[height = 0.113\textwidth]{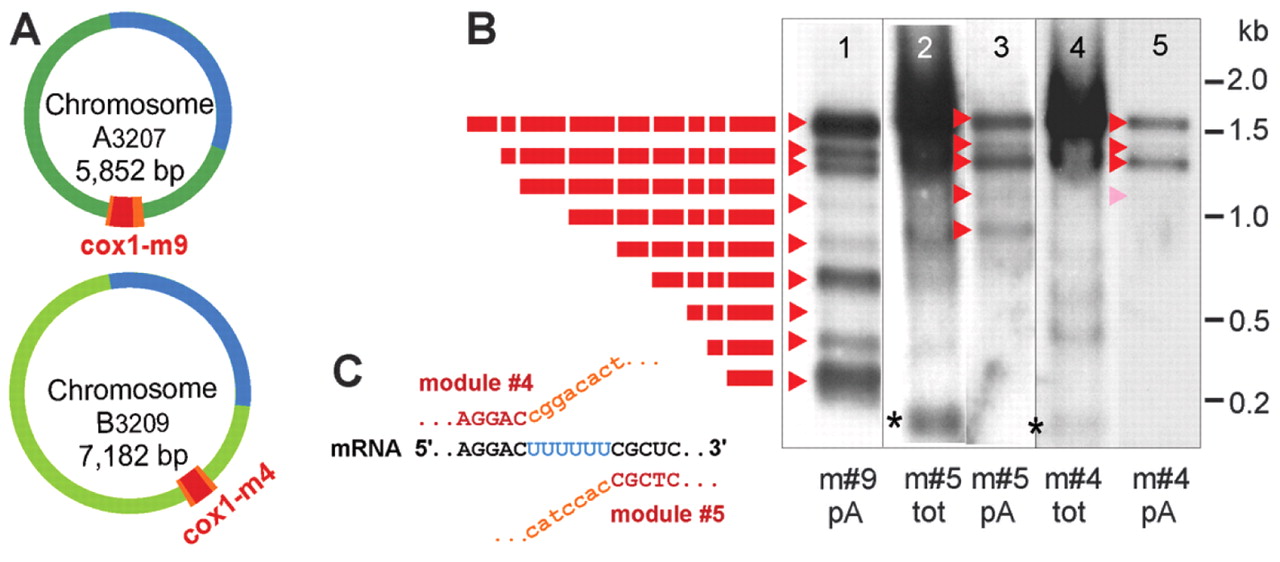}
\includegraphics[height = 0.113\textwidth]{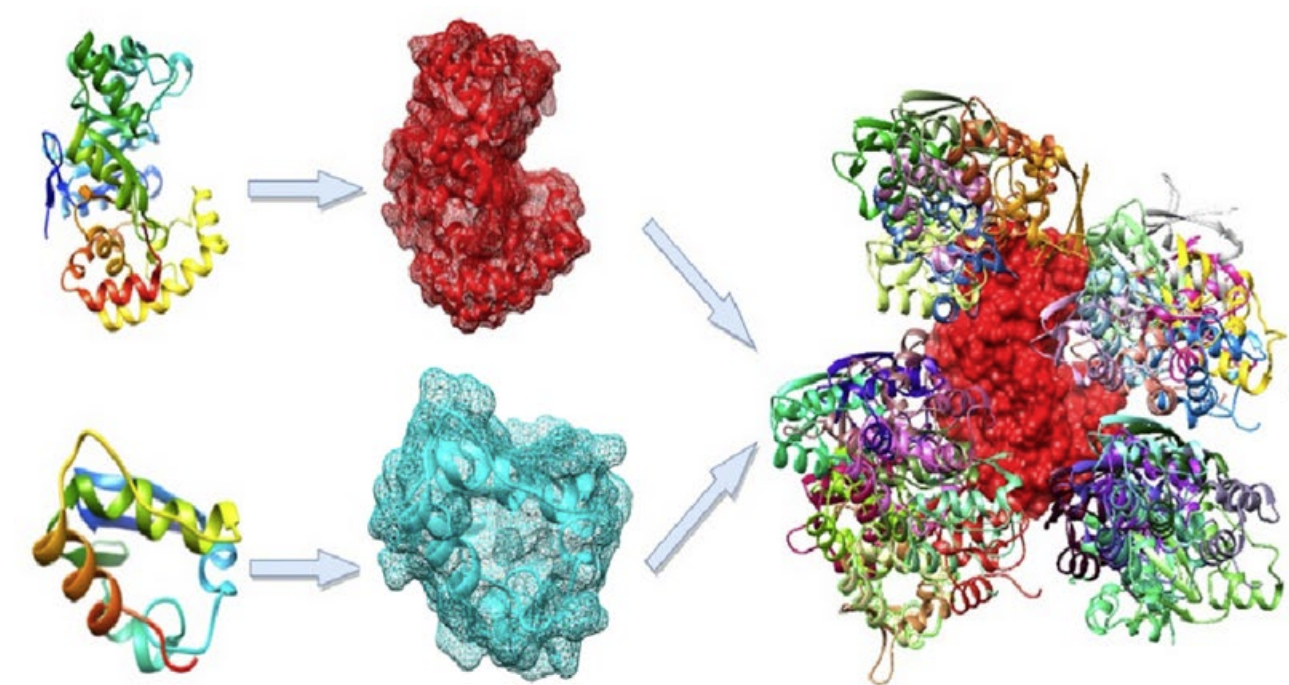}
\includegraphics[height = 0.144\textwidth]{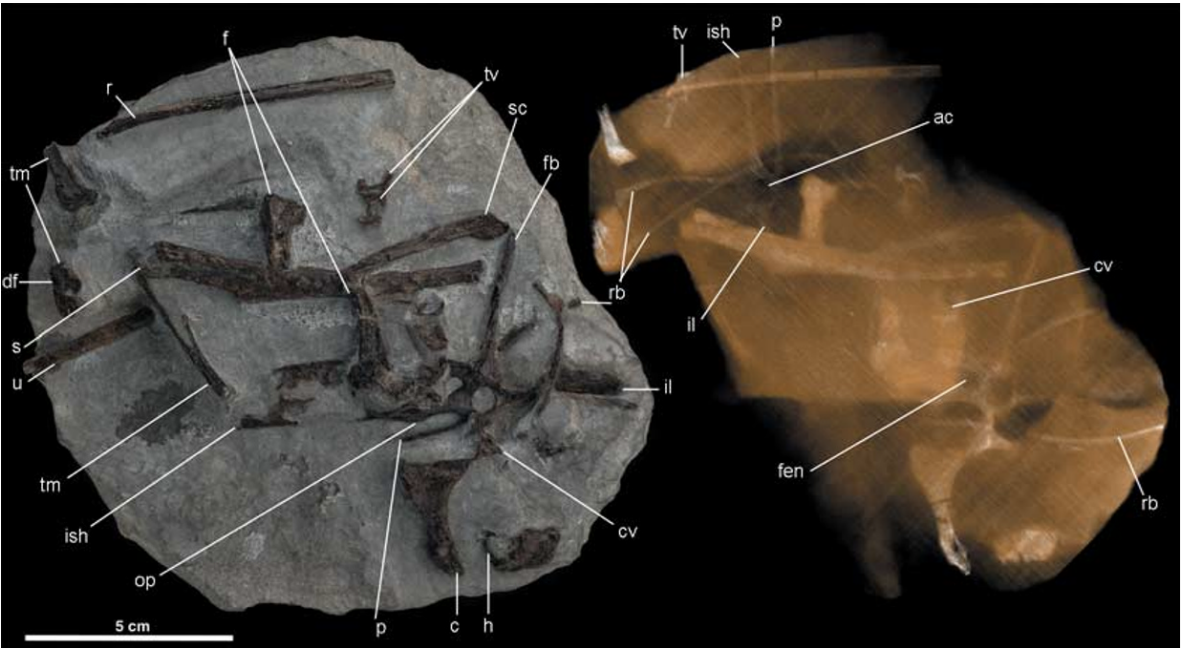}
\includegraphics[height = 0.144\textwidth]{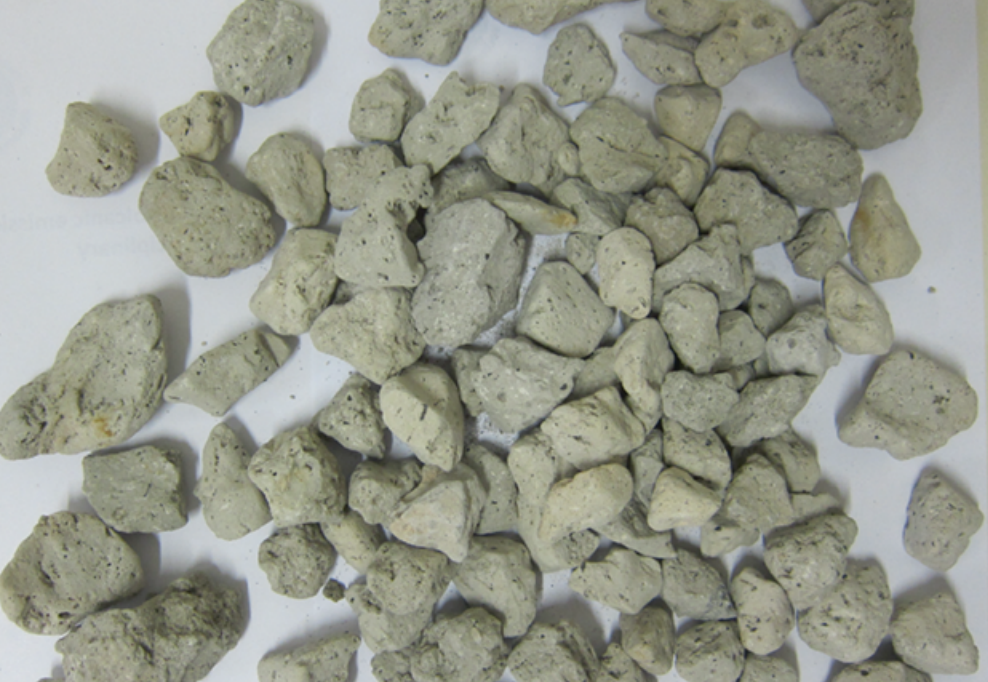}
\caption{Applications in scientific domains. (Top-Left) DNA sequencing~\cite{Science:DNA:2007} (Top-Right) Protein docking~\cite{PD:2022:Review} (Bottom-Left) Fossil reconstruction of a bird~\cite{Science:2005:Bird}. (Bottom-Right) Volcano crystal reconstruction~\cite{Volcano:Eruption:2020}.}
\label{Figure:Science:Apps}
\end{figure}

\subsection{Fractured Object Reassembly in Science}
\label{Subsec:Scientific:Applications}

There are many scientific applications for fractured object reassembly. One such example is docking of proteins and DNA Sequencing. In~\cite{Science:DNA:2007}, which mitochondrial genes from the unicellular eukaryote Diplonema are systematically fragmented into small pieces that are encoded on separate chromosomes and individually transcribed. The reassembly problem is to concatenate them into contiguous messenger RNA molecules. Another biological application is protein docking~\cite{PD:2014,PD:2022:Review}, where the goal is to find a rigid and non-rigid transformation of one 3D protein to interact with another protein.    

Another application is in paleontology, in which we can reconstruct fossils from fragments precisely in 3D, to allow 3D analysis of the whole shape. A semi-automatic approach was introduced in~\cite{Science:2005:Bird} to reconstruct an extinct bird. It is desirable to develop automatic approaches for more accurate and faster reconstructions. \cite{GUNZ200948} studied the virtual reconstruction of huminin crania to understand human evolution. \cite{10.3389/feart.2022.833379} introduced an interactive approach in Blender to restore a dorsal rib and frontal bone of a small-bodied Jurassic-age armored dinosaur from Africa, The digital restoration engaged all modalities of deformation (translation, rotation, scaling, distortion) and reconstruction (fracture infilling, adding missing bone, surface smoothing).

%Principles for the virtual reconstruction of hominin crania~\cite{GUNZ200948} We applied this method to  

One other potential scientific application is the reconstruction of crystals that ``blew up'' during eruption because of the fast decompression of trapped gases~\cite{Volcano:Eruption:2020}. In this setting, it is important to reconstruct not only the particles but also the voids within them that originally held the gases. Such reconstructions have great implications in geosciences.

\begin{figure}[b]
% \vspace{2in}
\includegraphics[width=\linewidth]{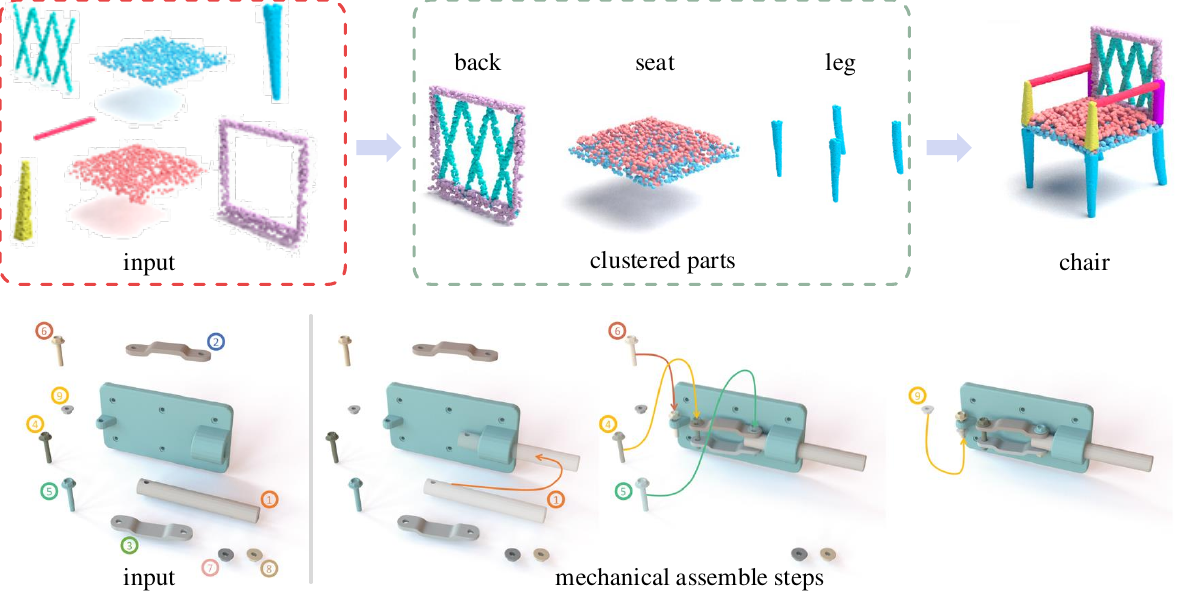}
\vspace{-0.1in}
\caption{Part Assembly examples. (Top) Object assembly~\cite{Du20243DHPA}. (Bottom) Mechanical assembly~\cite{AssembleThemAll}.}
\label{Figure:Part:Assembly}
\end{figure}

\begin{table*}[t]
\caption{Comparison of publicly available datasets for fractured object reassembly. Datasets are listed chronologically.}
\label{tab:datasets}
\centering
\adjustbox{max width=\linewidth}{
\begin{tabular}{l|ccccccc}
\toprule
Datasets & Objects/Groups & Pieces per Object & Total Pieces & Breakdown Type & Data Type & Data Origin & Ground Truth \\
\midrule
\href{https://www.geometrie.tuwien.ac.at/}{TU Wein}~\cite{10.1145/1141911.1141925} & 7 & 6-30 & 101 & Real & 3D & Artifacts & Roughly \\
\href{https://www.ntnu.edu/web/presious/presious}{Presious}~\cite{Theoharis_Papaioannou_2013} & 7 & 3-30 & 69 & Real & 3D & Cathedrals & Partial \\
\href{https://breaking-bad-dataset.github.io/}{BreakingBad}~\cite{DBLP:conf/nips/SellanCWGJ22} & 10474 & 2-100 & 8442044 & Synthetic & 3D & - & \cmark \\
\href{https://terascale-all-sensing-research-studio.github.io/FantasticBreaks/}{FantasticBreaks}~\cite{Lamb-2023-CVPR} & 150 & 2 & 300 & Real & 3D & Household &\cmark \\
\href{https://jiepengwang.github.io/FIRES/}{FIRES}~\cite{wang2023batch} & 15 & 7-18 & 123 & Real & 3D & Pottery & \cmark \\
\href{https://repairproject.github.io/RePAIR_dataset/}{RePAIR}~\cite{tsesmelis2024reassembling} & 18 & 2-44 & 1070 & Real & 2D/3D & Frescoes & Partial \\
\bottomrule
\end{tabular}
}
\end{table*}

\subsection{Part Assembly}

Part assembly refers to the process of assembling components that are decomposed in a semantically meaningful way, such as furniture components like table legs\cite{Mo_2019_CVPR} and mechanical parts like screws\cite{AssembleThemAll}. Although closely related to fracture assembly problem, part assembly introduces several key differences. It generally assumes that each part has an independent semantic function and that each component is complete and functional on its own, unlike fractured pieces in fracture assembly. Moreover, part assembly allows for duplicated components, and solutions are often non-unique. Unlike fracture assembly, where local or global geometry provides significant clues, part assembly relies more on semantic or functional understanding of the individual parts and the assembled object. However, despite these differences, we can identify valuable connections between the two problems, and both fields can benefit from exploring the shared foundations of these challenges.

A notable branch of part assembly research is the \textbf{object assembly} problem, which focuses on assembling parts to create complete daily objects such as tables, lamps, and chairs. Early work in this domain addressed two primary questions: how to collect relevant parts and how to combine them effectively. The foundational work in this area was proposed in Modeling by Example~\cite{modelingbyexample04}, where the authors introduced a method to retrieve parts from a repository of segmented 3D objects and combine them to create new objects from different parts. Building on this, SnapPaste~\cite{snappaste06} further developed the concept by focusing on how to align and model the mesh at the connecting areas of these parts. Several subsequent papers~\cite{ckgk-prabm-11, ckgk-prabm-11, assembly-based-pfg-16} introduced graphical probabilistic models to improve the selection and organization of retrieved parts. Additionally, research such as~\cite{data-driven-suggestions10, structure-recovery-part-12, ComplementMe17} explored retrieving or generating parts in an assembly process to complete objects from partial input, image data, or artist-created models. These developments laid the groundwork for later efforts to apply object assembly techniques to 3D asset generation~\cite{structurenet19, Componet_Schor_2019_ICCV, Li2019LearningPG}.

More recent work in object reassembly has shifted towards estimating the 6-DoF pose of input parts. Unlike earlier methods, these approaches typically assume that the provided parts are sufficient to form a complete object, similar to the process of assembling IKEA furniture. In support of this area, IKEA has released the IKEA 3D Assembly Dataset~\cite{Su2021IKEAObjectState}, which includes models and assembly manuals for six of their products. Additionally, the IKEA ASM Dataset~\cite{BenShabat2020IKEAASM} focuses on assembling IKEA objects using human demonstrations. On a broader scale, PartNet~\cite{Mo_2019_CVPR} has become a major benchmark for this research area. For example, Li et al.\cite{li2020impartass} used 2D images to guide the estimation of 3D poses for input parts. Techniques such as dynamic graph learning (DGL)~\cite{zhan2020generative} were proposed to generate part pose proposals, with later works incorporating RNNs~\cite{Narayan-2022-WACV} and transformers~\cite{Zhang2022IET, xu2024spaformersequential3dassembly} for improved accuracy. Researchers have also refined these methods by incorporating matching information~\cite{li2024category} and hierarchical structures~\cite{Du20243DHPA} to improve assembly precision.

Another key focus is the \textbf{mechanical assembly} problem, which differs from object assembly in that the provided parts are complete functional units, and the goal is to build a working mechanism. This problem often involves parts in the B-Rep format specific to mechanical engineering. Although this is a relatively new field, several papers have addressed different aspects of the problem. AutoMate~\cite{automate} presented an assembly-manual-based dataset of mechanical assemblies with 92,529 instances and proposed a graph convolutional network to predict mating locations and types for two input parts. JoinABLe~\cite{Willis2021JoinABLeLB} introduced an assembly dataset from the Fusion360GalleryDataset with 8,251 assemblies and 154,468 separate parts, along with a joint axis prediction network and a joint pose search algorithm based on predicted axes. Assemble Them All~\cite{AssembleThemAll}, which included 12,970 assemblies, addressed the challenge of multi-part, multi-scale input data with a disassembly-guided BFS algorithm.

Apart from these two areas, object assembly also emerged in LEGO assembly. \cite{wang2022translatingLego} proposed a segment and pose estimation network to translate the LEGO manual to executable planing, and StableLego~\cite{StableLego24} introduced an evaluation method to assess the stability of a LEGO assembly. 

Implicit in these assembly approaches are priors about the underlying complete objects and priors on how pairs or a subset of primitives interact. These two pairs are related to the global template shape priors and pairwise fragment matching in fractured object reassembly. As stated previously, the major difference is that part assembly focuses on semantic while fracture reassembly focuses on geometry. 
% \end{multicols}
% \begin{multicols}{2}

\begin{table*}[t]
\caption{Benchmark performance of published methods on the Breaking Bad Dataset. All approaches were initially trained and tested on the \texttt{everyday} object subset. They are then fine-tuned and tested on the \texttt{artifact} subset (unless noted with b). Testing was conducted on objects containing 2-20 fragments. Methods are listed chronologically.}
\label{tab:benchmark_result}
\begin{threeparttable}
\centering
% \resizebox{\linewidth}{!}{
\adjustbox{max width=\linewidth}{
\begin{tabular}{l|ccccccc|cccccc}
\toprule
                & \multicolumn{7}{c|}{\texttt{everyday}}                              & \multicolumn{6}{c}{\texttt{artifact}}                      \\
\midrule
                & RMSE(R) & MAE(R) & RMSE(T) & MAE(T) & CD   & PA   & time   & RMSE(R) & MAE(R) & RMSE(T) & MAE(T) & CD   & PA   \\
                & degree  & degree & $\times 10^{-2}$ & $\times 10^{-2}$ & $\times 10^{-3}$ & \% & ms & degree  & degree & $\times 10^{-2}$ & $\times 10^{-2}$ & $\times 10^{-3}$ & \% \\
\midrule
\href{https://github.com/Wuziyi616/multi_part_assembly}{Global}~\cite{Li2019LearningPG}          & 80.7    & 69.7   & 14.8    & 11.8   & 14.6 & 21.8 & 23.7   & 83.8    & 71.8   & 16.6    & 13.8   & 19.0 & 13.3 \\
\href{https://github.com/Wuziyi616/multi_part_assembly}{LSTM}~\cite{9157694}            & 84.2    & 72.7   & 16.2    & 12.7   & 15.8 & 19.4 & 29.9   & 84.6    & 73.1   & 16.8    & 14.0   & 21.5 & 11.7 \\
\href{https://github.com/Wuziyi616/multi_part_assembly}{DGL}~\cite{zhan2020generative}             & 79.4    & 67.8   & 15.8    & 12.5   & 14.3 & 23.9 & 28.7   & 81.7    & 69.7   & 16.6    & 13.8   & 17.3 & 19.4 \\
\href{https://github.com/crtie/Leveraging-SE-3-Equivariance-for-Learning-3D-Geometric-Shape-Assembly}{SE(3)-Equiv}~\cite{Wu_2023_ICCV}\tnote{c}     & 79.3    & -      & 16.9    & -      & 28.5 & 8.4  & 129.9  & -       & -      & -       & -      & -    & -    \\
\href{https://github.com/Jiaxin-Lu/Jigsaw}{Jigsaw}~\cite{lu2023jigsaw}          & 42.3    & 36.3   & 10.7    & 8.7    & 13.3 & 57.3 & 1063.5 & 52.4    & 45.4   & 22.2    & 19.3   & 14.5 & 45.6 \tnote{b} \\
\href{https://github.com/521piglet/PHFormer}{PHFormer}~\cite{DBLP:conf/aaai/CuiYD24}        & 26.1    & 22.4   & 9.3     & 7.5    & 9.6  & 50.7 & 10.7   & 32.1    & 27.7   & 11.4    & 9.5    & 11.8 & 37.1 \\
\href{https://github.com/IIT-PAVIS/DiffAssemble}{DiffAssemble}~\cite{Scarpellini-2024-CVPR}    & 73.3    & -      & 14.8    & -      &      & 27.5 & -      & -       & -      & -       & -      & -    & -    \\
\href{https://github.com/eric-zqwang/puzzlefusion-plusplus}{PuzzleFussion}++~\cite{wang2024puzzlefusionautoagglomerative3dfracture} & 38.1    & -      & 8.0     & -      & 6.0  & 70.6 & 928.5  & 52.1    & -      & 13.9    & -      & 14.3 & 49.6 \tnote{b} \\
\midrule
\href{https://github.com/NahyukLEE/pmtr}{PMTR}~\cite{lee20243dgeometricshapeassembly} \tnote{a}    & 31.6    & -      & 9.9     & -      & 5.6  & 70.6 & -      & 31.6    & -      & 10.1    & -      & 4.3  & 71.6 \\
\href{https://github.com/xuqunce/FragDiff}{FragmentDiff}~\cite{fragmentdiff} \tnote{a}    & 13.7    & 11.5   & 7.4     & 5.8    & -    & 90.2 & -      & 18.2    & -      & 8.1     & -      & -    & 82.3 \tnote{b} \\
\bottomrule
\end{tabular}}
\begin{tablenotes}
    \item[a] Reported model is trained and tested on the ``volume-constrained'' subset.
    \item[b] Reported model is only trained on \texttt{everyday} subset and is not finetuned on the \texttt{artifact} subset.
    \item[c] Original paper tested on 2-8 fragments; results for 2-20 fragments are cited from PuzzleFusion++.
\end{tablenotes}
\end{threeparttable}
\vspace{-0.2in}
\end{table*}

\begin{figure}[t]
\centering
\includegraphics[width=0.505\linewidth]{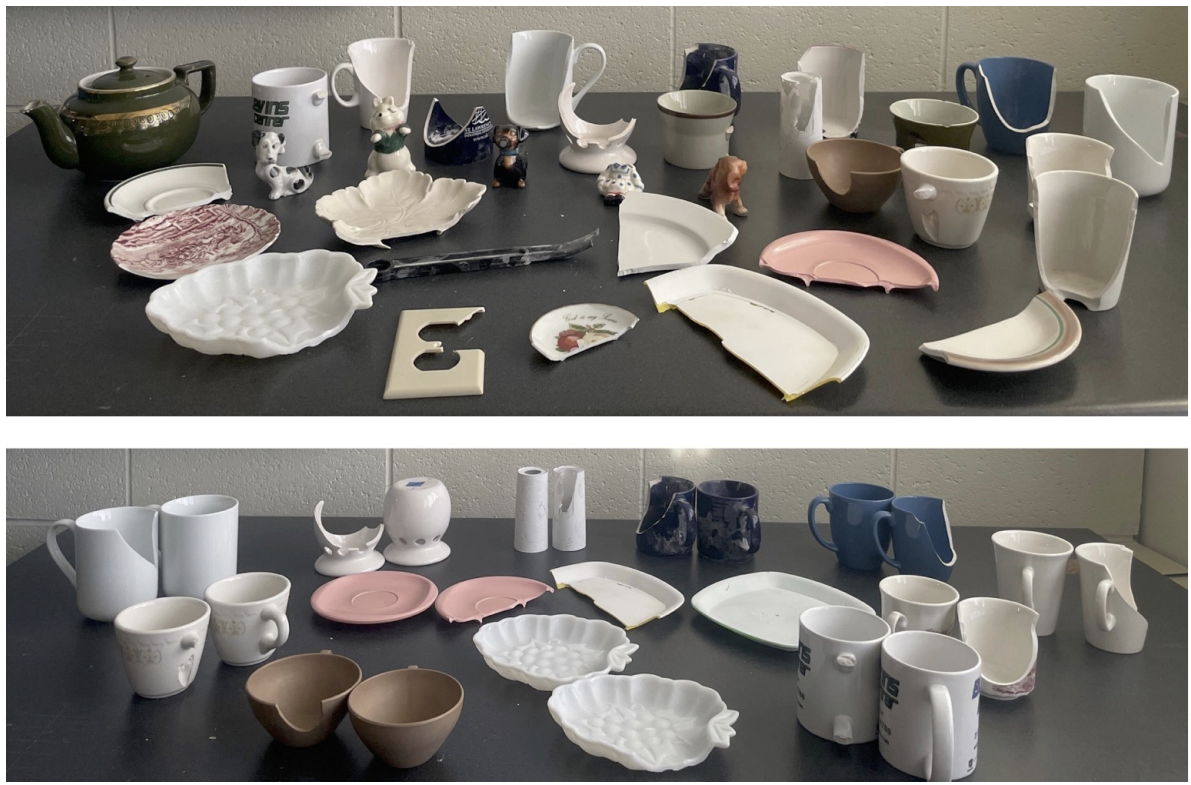}
\includegraphics[width=0.485\linewidth]{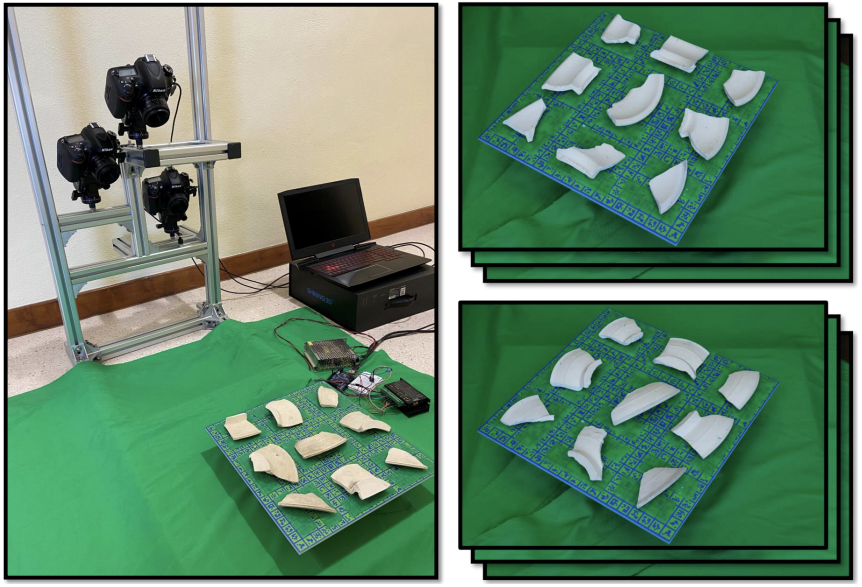}
\includegraphics[trim={3cm 0.8mm 2.5cm 0},clip,width=0.505\linewidth]{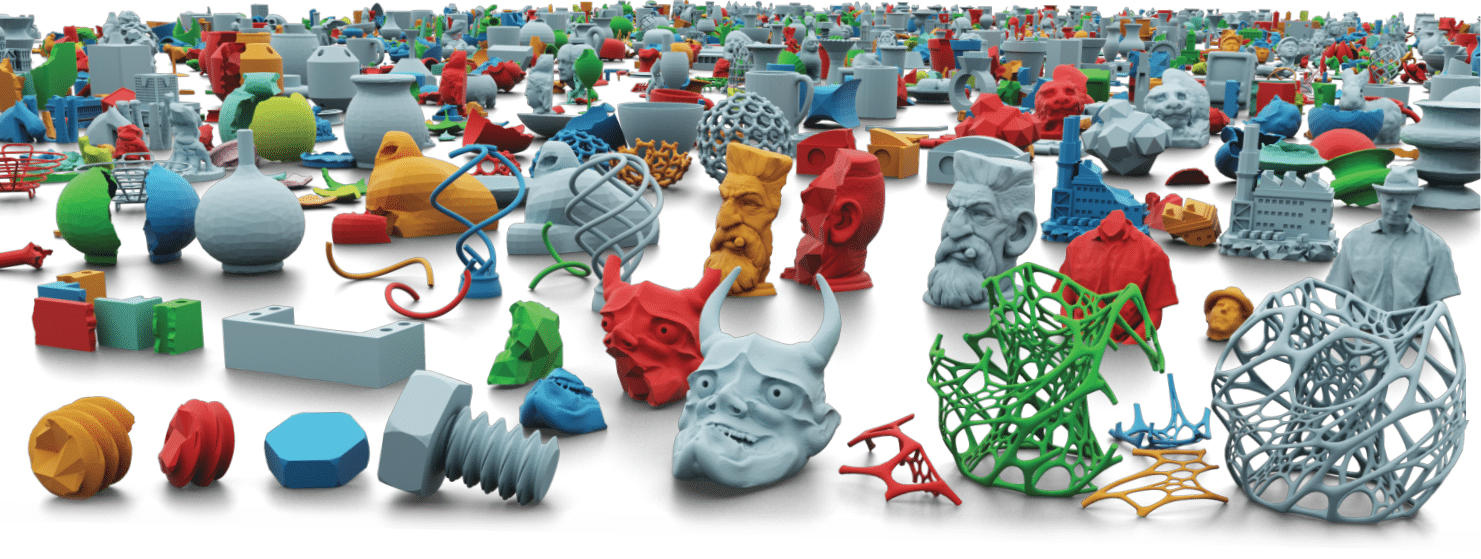}
% \hspace{0.1mm}
% \includegraphics[trim={0 0 5cm 0},clip,width=0.485\textwidth]{figs/Sec8Datasets/repair_dataset.jpg}
\includegraphics[trim={0 0.1cm 0 0},clip,width=0.475\linewidth]{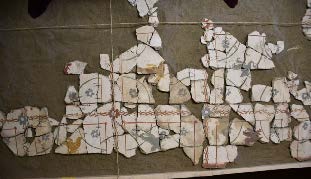}
% \vspace{-0.1in}
\caption{Datasets used for deep learning research in fractured object reassembly. (Top-Left) Fantastic Breaks dataset~\cite{Lamb-2023-CVPR} which combines both isolated fragments (top) and paired complete objects and their fragments (bottom). (Top-Right) A real dataset of fragments with ceramic material~\cite{wang2023batch} is captured using the system shown here. (Bottom-Left) Synthetic dataset created by performing fracture simulation on a diverse set of objects~\cite{DBLP:conf/nips/SellanCWGJ22}. (Bottom-Right) RePAIR dataset~\cite{tsesmelis2024reassembling} which contains scans of a fresco at Pompeii archaeological park and ground truth provided by archaeologists.}
\vspace{-0.2in}
\label{Figure:Datasets}    
\end{figure}

\section{Datasets and Open Source} 
\label{Section:Datasets:Software:Packages}

We first discuss existing datasets in Section~\ref{Subsec:Datasets} and the evaluation metrics used in benchmarks in Section~\ref{Subsec:Evaluation:Metrics}. We then discuss existing open-sourced methods in Section~\ref{Subsec:Software:Packages}.

\subsection{Datasets}
\label{Subsec:Datasets}

Machine learning approaches have shown great advancements in the task of reassembly of fragments, but are highly dependent on large-scale datasets. However, the first data sets used in fragment reassembly research are primarily small-scale, often manually created \cite{10.1145/1141911.1141925, Brown:2012:TFV, Stanford:2006:2,Theoharis_Papaioannou_2013} or derived from automated semantic segmentation \cite{automate, Mo_2019_CVPR}.

% \cite{Mo_2019_CVPR} \cite{stanfordFUR} \cite{Brown:2012:TFV} \cite{automate} \cite{10.1145/1141911.1141925}

Several recent efforts have focused on scanning fragmented objects at scale. The first example is Fantastic breaks~\cite{Lamb-2023-CVPR}, which includes a small collection of objects with broken fragments and more objects that were broken into fragments by the authors (see Figure~\ref{Figure:Datasets}(Top-Left)). Both fragments and complete objects (if they exist) are scanned. \cite{wang2023batch} developed a system to batch scan fragments of objects with ceramic material (see Figure~\ref{Figure:Datasets}(Top-Right)). This dataset has ground-truth complete objects for evaluation. In addition, the scale of these two datasets makes it possible to develop deep neural networks for reassembly. However, the number of categories in these datasets is still small, presenting generalization issues in novel categories. The RePAIR dataset~\cite{tsesmelis2024reassembling} (see Figure~\ref{Figure:Datasets}(Bottom-Right)) represents a new advance in real-world fragment assembly datasets, built from a collapsed fresco excavated from archaeological sites. It provides both 2D and 3D puzzles along with high-resolution texture information and expert archaeological annotations. It includes scenarios with missing pieces, highly irregular fragment shapes, and unknown matching relationships between pieces, which breaks a key assumption present in most existing object-level assembly datasets - that all pieces originate from a single, complete object.

In addition to real data sets, Breaking Bad~\cite{DBLP:conf/nips/SellanCWGJ22} is a synthetic data developed by running a simple fracture simulation procedure to generate fragments of shape collections that undergo large geometric and topological variations (see Figure~\ref{Figure:Datasets}(Bottom-Right)). This dataset is interesting for fractured object reassembly because the underlying objects belong to diverse object categories. For example, template-based reassembling techniques can be developed by learning a shape prior of the complete objects in this dataset. Just as the role of ShapeNet~\cite{DBLP:journals/corr/ChangFGHHLSSSSX15} in boosting 3D deep learning rich, we envision that research progress on algorithmic reassembly of fractured objects will be accelerated with Breaking Bad and similar datasets. However, the fragments are generated using non-physics based fracture simulation. The question is how the reassembly pipelines derived from such synthetic datasets generalize to real data. We present a comparison of publicly available datasets in Table~\ref{tab:datasets}.

\begin{figure*}[th]
% \vspace{2in}
\includegraphics[width=\textwidth]{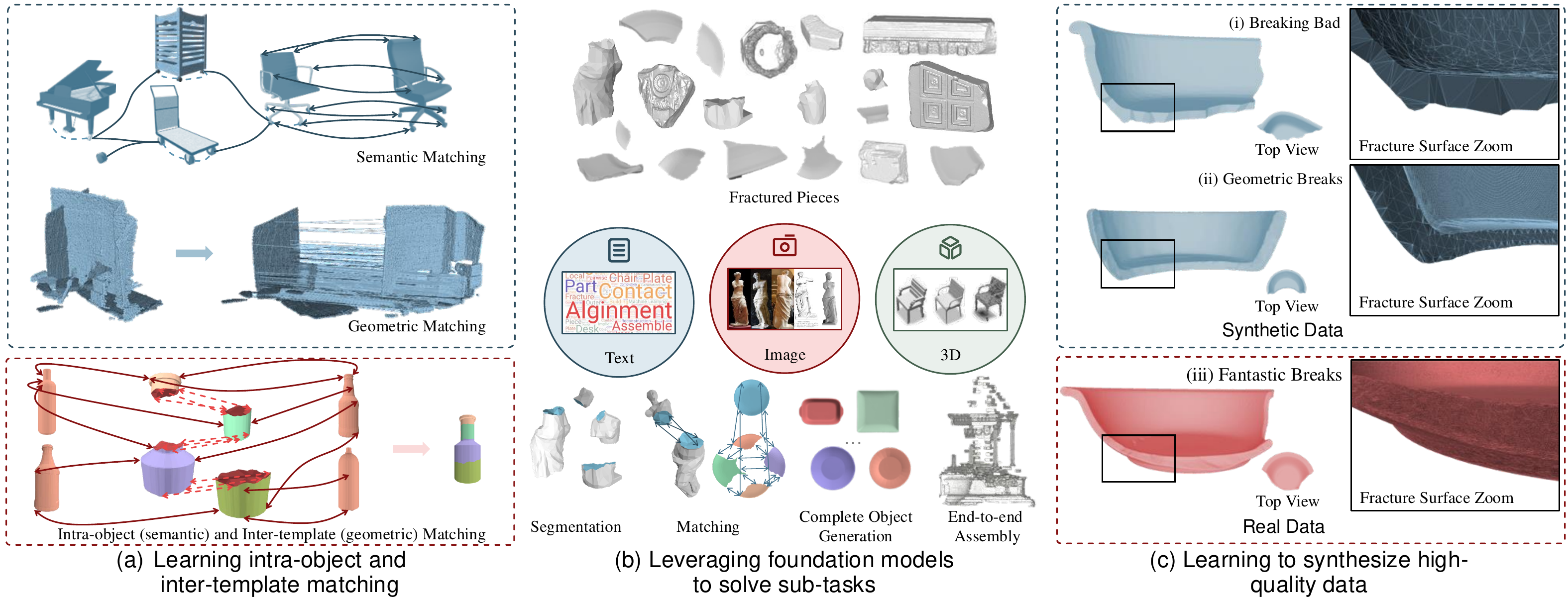}
% \vspace{-0.15in}
\caption{Illustrations of future directions (Left) Representations and algorithms for matching complete template shapes with large geometric and topological variations and end-to-end learning of the entire reassembly pipeline. (Middle) Leveraging foundation models to solve sub-tasks such as segmenting fractured surfaces, matching fragments, predicting complete template shapes from input fragments, and matching complete template shapes and fragments. (Right) Learning to generate synthetic datasets of 3D fragments and doing so jointly with learning the reassembly network.}
% \vspace{-0.25in}
\label{Figure:Future:Directions}    
\end{figure*}

\subsection{Evaluation Metrics}
\label{Subsec:Evaluation:Metrics}
To quantitatively evaluate reassembly methods, the research community has adopted several standard metrics that measure both the accuracy of predicted transformations and the quality of the final assembly.

For rotation evaluation, two primary metrics are commonly used: the Mean Absolute Error (MAE) and Root Mean Square Error (RMSE). Given a predicted rotation $\tilde R$ and ground-truth rotation $R^{gt}$ represented in Euler angles, these metrics are defined as:
\begin{equation}
    \textup {MAE(R)} = \frac{1}{3} ||\tilde R- R^{gt}||_1, \quad
\textup {RMSE(R)} = \frac{1}{\sqrt 3} ||\tilde R-R^{gt}||_2
\end{equation}
Similarly, for translation evaluation, given a predicted translation $\tilde{\bs{t}}$ and ground-truth translation $\bs{t}^{gt}$, the corresponding metrics are:
\begin{equation}
\textup {MAE(T)} = \frac{1}{3} ||\tilde {\bs t}- \bs{t}^{gt}||_1,
\quad
\textup {RMSE(T)} = \frac{1}{\sqrt 3} ||\tilde {\bs t}-\bs{t}^{gt}||_2
\end{equation}
To evaluate the quality of the assembled result, two additional metrics are widely used. Chamfer Distance (CD) evaluates the overall assembly quality by measuring the bi-directional point cloud distance between the assembled result and the ground truth shape:
\begin{equation}
    \textup{CD}(X,Y) = \frac{1}{|X|}\sum_{x\in X}\min_{y\in Y}||x-y||_2 + \frac{1}{|Y|}\sum_{y\in Y}\min_{x\in X}||y-x||_2
\end{equation}
Part Accuracy (PA) measures the percentage of correctly placed fragments. A fragment is considered correctly placed if its Chamfer distance to the ground truth is below a threshold (typically 0.01):
\begin{equation}
\textup{PA} = \frac{\text{number of fragments with CD} < 0.01}{\text{total number of fragments}}
\end{equation}
Finally, computational efficiency is measured through the total execution time of the algorithm, typically measured in milliseconds. This metric is particularly important for practical applications where real-time or near-real-time performance is desired.

For multi-piece reassembly tasks, the transformation errors (MAE and RMSE) and CD are typically computed as the mean error across all constituent pieces of each object. The overall performance of a method is then evaluated by averaging these object-level metrics across the entire test set. While these metrics provide a standardized way to compare different reassembly approaches, it is worth noting that they may not fully capture aspects such as physical validity or aesthetic quality of the final assembly.

\subsection{Open Source}
\label{Subsec:Software:Packages}
%\qixing{Jiaxin}

Few early approaches to fractured object reassembly released code. A major factor is that in the early days, we did not have established benchmark datasets for experimental evaluation. For example, there are rich benchmark datasets in computer vision and researchers deliver open-source codes that execute on benchmark datasets. With the introduction of large-scale benchmark datasets, we see the trend to release codes. For example, many papers that evaluate using Breaking Bad have open-source implementations, including Jigsaw~\cite{lu2023jigsaw}, SE(3) equivariance~\cite{Wu_2023_ICCV}, PHFormer~\cite{DBLP:conf/aaai/CuiYD24}, DiffAssemble~\cite{Scarpellini-2024-CVPR}, PMTR~\cite{lee20243dgeometricshapeassembly}, and PuzzleFusion++~\cite{wang2024puzzlefusionautoagglomerative3dfracture}. We provide a comprehensive benchmark performance on these open-sourced methods in Table~\ref{tab:benchmark_result}. For procedural methods such as ICP on the large-scale dataset, we encourage the readers to refer to the results presented in Neural Shape Mating~\cite{Chen_2022_CVPR}.

% \end{multicols}
% \begin{multicols}{2}
\section{Future Directions}
\label{Section:Future:Directions}

We are in the deep learning era, in which there are breakthrough results after breakthrough results on developing deep learning approaches to solve core problems in geometry processing and 3D vision. Undoubtedly, the future of tackling the reassembly problem lies in deep learning approaches. Although we have discussed many recent advances on this front, there are open questions from neural representations to algorithms to training data, which should be addressed before deploying reassembly pipelines in practice.

On the representation and algorithm side, there are open questions on how to learn shape priors of diverse complete objects. When targeting general object categories that have large topological and geometrical variations, implicit shape representations, which can encode shapes with varying topology, become the only choice. However, establishing reliable correspondences between implicit representations of complete shapes and the underlying geometric structure of fragments remains an open research challenge.

Once we have addressed this open problem, it makes sense to study how to integrate all components of the reassembly pipeline in an end-to-end manner, enabling effective training. Currently, it remains an open question on how to do so when the reassembly pipeline uses deforming template shapes. 

In addition, having sufficient training data is critical to the success of developing deep neural networks. However, this becomes challenging in the case of the fragment reassembly problem. First of all, the number of complete objects that have fragments remains sparse. In addition, developing software and hardware systems to scan these fragments is very costly. There are two interesting topics related to addressing this data issue. 

The first topic explores boosting reassembly performance using vision foundation models. Many subtasks in the reassembly pipeline - including fracture surface segmentation, establishing correspondences between pairs of fragments, predicting the underlying complete object from fragments, and matching fragments with a complete object - have counterparts in core geometry processing and 3D vision problems. However, these tasks face unique challenges in the reassembly context. For correspondence establishment, current methods primarily rely on geometric matching using surface features or contours, which shows limited generalization ability across different object categories and fracture patterns. The assembly problem requires both geometric matching between fragments and semantic matching between fragments and templates - a dual objective that few existing approaches address. Additionally, while there has been significant progress in 3D shape generation using implicit representations and neural fields, predicting complete objects from fragments remains challenging. Most point cloud generation methods struggle with the diversity of possible completions, especially when handling rotated or partially observed objects. Recently, we have seen that 2D foundation models, either through rendered images or lifted 2D features, have led to breakthrough results on these tasks. 2D pretrained model, e.g. DINOv2~\cite{oquab2024dinov} and Stable Diffusion~\cite{stablediffusion}, have shown potential as depth estimation backbone. The extracted features have also shown success in finding correspondences~\cite{zhang2023tale,tang2023emergent,hedlin2023unsupervised} with a 30\% improvement. This brings interesting questions on how to adopt these approaches in the corresponding subtasks in reassembly. For example, a recent paper~\cite{Zhai-2024-CVPR} has shown that vision-language models can effectively understand the material properties of objects from rendered images. However, rendering fragments presents unique challenges compared to complete objects, as fracture regions need special consideration for effective visual understanding. Developing rendering techniques that can reveal and emphasize fracture characteristics while maintaining other geometric and material properties remains an open research question.

The second topic is on generating synthetic fracture data to train fracture reassembly networks. The advantage of this approach is that we have abundant complete 3D objects, allowing us to learn shape priors. Moreover, we have ground-truth poses and correspondences from synthetic data, which are hard to obtain from real data. In fact, several synthetic datasets~\cite{DBLP:conf/nips/SellanCWGJ22} have been introduced in the past few years. As discussed in this survey, these datasets have led to a wave of research activities on developing deep learning pipelines to solve the reassembly problem. However, a fundamental challenge when applying these deep reassembly pipelines in practice is the domain gap between synthetic data and real data. To this end, it is important to develop an end-to-end learnable fracture simulation pipeline, so that we can control the characteristics of the simulation output, e.g., the number of fragments and erosion among fracture surfaces. This network should be physics-aware, so that we do not need to generate a lot of accurate but costly training data for supervision. This network should also properly model the weathering effects, which are prevalent in real datasets. Such a network allows us to learn hyper-parameters of fracture simulation, e.g., material properties and distribution of external forces, from simulated data to maximize its performance on real data.  

Once we have an end-to-end learnable fracture simulation pipeline, another important problem is studying how to combine fracture reassembly and simulation in an end-to-end manner. This allows optimizing the simulation network so the reassembly network generalizes optimally when trained from simulation output. This task goes beyond standard end-to-end training, as we must develop losses on the output distribution of the simulation network.

From a community-building perspective, it is important to create large-scale benchmark datasets for experimental evaluation. This will attract more researchers to work on this problem and, according to the trend in computer vision, will stimulate more researchers to release open-source codes. We do observe a good initial trend from the Breaking Bad dataset~\cite{DBLP:conf/nips/SellanCWGJ22}.

Finally, it is also important to identify more applications of fractured object reassembly in scientific domains. This will amplify the broader impacts of this research community. 

% \end{multicols}
\section{Conclusions}
\label{Section:Conclusions}

In this survey, we have provided a thorough survey of the reassembly problem, which finds rich applications in cultural heritage restoration, bone fracture restoration, and object/scene modeling from parts and objects. We have covered early procedural approaches to more recent deep learning based approaches. Due to connections to core geometry processing problems in shape segmentation, shape matching, symmetry and primitive detection, and learning shape priors, we also discussed relevant approaches to these problems, emphasizing similarities and differences. We also covered current state-of-the-art methods for solving real problems in these application domains. 

We envision that this topic will undergo significant progress in the coming years from the perspective of developing deep learning based approaches. We have highlighted two promising venues for future research. The first studies how to leverage foundation models to boost various sub-problems of reassembly. The second examines how to generate synthetic data in a controlled manner, which enables joint learning of reassembly and fracture simulation. We conclude that this very fundamental research field can also benefit from having more scientific applications. 

% \textbf{Acknowledgement.} Qixing Huang is supported by NSF Career IIS-2047677, NSF IIS-2413161, and Gifts from Adobe and Google. Xin Li is partly supported by NSF CBET-2115405 and NIH R15HD108765. Junfeng Jiang would like to acknowledge support from Jiangsu Province Key Research Program-Social Development-Clinical Frontier Technologies (BE2022718).
% Datasets introduced in the early days are associated witht the corresponding articles. Stanford project (The Forma Urbis Romae Fragment Database), Qixing's SIG06 paper, check Pinceton's efforts.

\section*{Acknowledgement} Qixing Huang is supported by NSF Career IIS-2047677, NSF IIS-2413161, and Gifts from Adobe and Google. Xin Li is partly supported by NSF-IIS 1320959 and NSF CBET-2115405. Junfeng Jiang would like to acknowledge support from Jiangsu Province Key Research Program-Social Development-Clinical Frontier Technologies (BE2022718).

\bibliographystyle{unsrt}  
% \bibliography{egbib.bib}  %%% Remove comment to use the external .bib file (using bibtex).



% George Kour and Raid Saabne.
% \newblock Real-time segmentation of on-line handwritten arabic script.
% \newblock In {\em Frontiers in Handwriting Recognition (ICFHR), 2014 14th
%   International Conference on}, pages 417--422. IEEE, 2014.

% George Kour and Raid Saabne.
% \newblock Fast classification of handwritten on-line arabic characters.
% \newblock In {\em Soft Computing and Pattern Recognition (SoCPaR), 2014 6th
%   International Conference of}, pages 312--318. IEEE, 2014.

% \bibitem{hadash2018estimate}
% Guy Hadash, Einat Kermany, Boaz Carmeli, Ofer Lavi, George Kour, and Alon
% \newblock Estimate and replace: A novel approach to integrating deep neural
%   networks with existing applications.
% \newblock {\em arXiv preprint arXiv:1804.09028}, 2018.

% \end{thebibliography}

\end{document}